**Multilingual Large Language Models do not comprehend all natural languages to equal degrees**

Natalia Moskvina[1], Raquel Montero[1], Masaya Yoshida[1,2], Ferdy Hubers[3], Paolo Morosi[1], Walid Irhaymi[1], Jin Yan[1], Tamara Serrano[1], Elena Pagliarini[1], Fritz Günther[4] & Evelina Leivada[1,2]

1. Universitat Autònoma de Barcelona
2. Institució Catalana de Recerca i Estudis Avançats (ICREA)
3. Radboud University Nijmegen
4. Humboldt-Universität zu Berlin

Large Language Models (LLMs) play a critical role in how humans access information. While their core use relies on comprehending written requests, our understanding of this ability is currently limited, because most benchmarks evaluate LLMs in high-resource languages predominantly spoken by Western, Educated, Industrialised, Rich, and Democratic (WEIRD) communities. The default assumption is that English is the best-performing language for LLMs, while smaller, low-resource languages are linked to less reliable outputs, even in multilingual, state-of-the-art models. To track variation in the comprehension abilities of LLMs, we prompt 3 popular models on a language comprehension task across 12 languages, representing the Indo-European, Afro-Asiatic, Turkic, Sino-Tibetan, and Japonic language families. Our results suggest that the models exhibit remarkable linguistic accuracy across typologically diverse languages, yet they fall behind human baselines in all of them, albeit to different degrees. Contrary to what was expected, English is not the best-performing language, as it was systematically outperformed by several Romance languages, even lower-resource ones. We frame the results by discussing the role of several factors that drive LLM performance, such as tokenization, language distance from Spanish and English, size of training data, and data origin in high- vs. low-resource languages and WEIRD vs. non-WEIRD communities.

# Introduction

Recent advances in artificial intelligence (AI) systems have expanded their presence in everyday life to an unprecedented degree, with an increasing range of tasks now delegated to AI assistants by individual users and large organisations alike. Reliance on such tools, particularly the field's most influential systems, Large Language Models (LLMs), stems from their strong ability to generate tailored responses to virtually any query, as well as their accurate performance on various benchmarks across knowledge domains.[1-6] These outputs are largely achieved through learning shortcuts that are based on the extraction of statistical patterns from natural language; a fact that has generated significant interest in cognitive science and linguistics, prompting the idea that LLMs could be considered computational models of human language and cognition.[7-9]

At the same time, a growing number of studies have expressed reservations about the true extent of the linguistic abilities that are underlying LLMs' remarkable language performance in several tasks. While models seem to master certain aspects of language reasonably well (e.g., long-distance number agreement, [10] various types of constructions[11,12], polysemy patterns[13]), they may fail to reach human-like baselines in a variety of other linguistic domains (e.g., pragmatic implicatures[14], quantification[15], generic expressions[16], passive voice and negation[17], and language comprehension[18]). This distinct behaviour indicates that the underlying language system of LLMs is different from that of humans: the gaps in their linguistic abilities often stem from aspects of their architecture and learning process,[19] and although these shortcomings might seem trivial at first glance, they can have tangible consequences, particularly in high-stake applications where even minor misinterpretation can be harmful[20]. Relatedly, LLM limitations have mostly been reported for English or other major Indo-European languages that have a vast amount of linguistic data available for model training (i.e., the so-called *high-resource languages*). This raises the possibility that even more pronounced deviations may emerge when models are asked to perform in languages spoken by smaller linguistic communities (i.e., *low-resource languages*). Indeed, the distribution of data across languages in LLM training is highly uneven (also possibly in terms of including more noisy or biased training data), reflecting a broader bias towards over-representing White, Educated, Industrialised, Rich, and Democratic (WEIRD) communities in cognitive and behavioural research, to the detriment of the visibility of other groups that often remain in the margins.[21] Since WEIRD populations represent a small and potentially privileged portion of humanity, it has long been recognised that focusing nearly exclusively on these groups in experimental testing fails to provide a comprehensive and truly diverse representation of human cognition.[22,23] Yet, it remains unexplored to what extent this bias is mirrored in the performance of AI systems and the conclusions we draw about them, with training pipelines and evaluation practices heavily skewed towards English and a handful of other high-resource Western languages.[24] While notable exceptions exist —for example, Chinese has received considerable attention in AI research[25] — the overall trend is concerning, as it can result in imbalanced knowledge representation and may limit fair cross-linguistic access to reliable AI tools.

Given the limitations in our knowledge of how LLMs perform in different languages, it is important to systematically assess their cross-linguistic performance using a uniform metric. Although modern commercial models are trained on a broad range of languages and are reported to exhibit strong multilingual capabilities, [26-29] it is not clear how their internal multilingual space is organised. In particular, we do not yet know how well typologically diverse, genetically unrelated, high- vs. low-resource languages are captured by LLMs and whether different kinds of models are better in capturing some languages vs. others.

Previous research has highlighted imbalances in models' overall performance across languages,[30-32] and stronger outcomes have been predictably observed for English and other high-resource, Latin-based scripts languages.[33-37] In an attempt to explain this variation, Zhang et al.[37] adapt cognitive models of human bilingualism and argue that GPT-3.5 exhibits signs of subordinate multilingual organisation, with English as the dominant language in terms of the model's internal representation, and outputs in other languages being a product of an internal translation process, rather than direct prompt-response generation. It has been further hypothesised that these cross-linguistic disparities may stem from limitations related to current data collection methods and model-training techniques, thus possibly not fully fixable with an increase in data quality and quantity.[37] However, this analysis relies heavily on datasets generated by the very model under examination, which limits the interpretability and generalisability of the claims.

In other work, the superiority of English is generally taken as a premise and treated as an unquestioned baseline. For example, Xu et al.[38] propose a method for identifying cross-linguistic weaknesses in LLMs by modifying prompts to amplify performance differences between English and other target languages. Their analysis suggests that linguistic proximity may explain certain shared weaknesses, although proximity is not defined by an established metric, and languages are instead grouped into broad categories of Asian and European language families. Similarly, Wang et al.[39] report greater consistency in performance and mutual information dissemination in high-resource, typologically related languages. However, their analysis is limited to correlations of varying strength and would thus benefit from more rigorous statistical treatment.

Nonetheless, English dominance is not completely undisputed, with some studies reporting cases in which other languages outperform English. For instance, an analysis of GPT's ability to generalise morphological patterns across English, German, Turkish, and Tamil reveals highest accuracy in German.[40] In a larger study of 26 languages, Kim et al.[41] found that Romance languages, along Polish and Russian, overall outperform English in long-context retrieval tasks, and when controlling for amount of input rather than number of tokens, Slavic languages surpass Romance ones. These results point to potential tokenization effects and suggest that English may not always be the strongest-performing language, contrary to widespread assumptions.

Taken together, previous work demonstrates that LLMs exhibit substantial cross-linguistic disparities, with speakers of major Indo-European, Latin-based script languages likely having access to more reliable AI tools. However, most of the relevant studies have focused almost exclusively on multilingual variation in information handling, while largely overlooking the root of the issue: How language itself is processed by models. Since LLMs rely on human-curated linguistic datasets for building their internal representations, variation in their performance in different languages must ultimately reflect differences in how they comprehend and/or capture structures particular to each language. This calls for evaluation methods grounded in linguistic theory and based on benchmarks designed by human experts rather than datasets produced through machine translation.

The present study addresses this gap by evaluating LLMs cross-linguistically on the very task they are expected to perform best, namely language comprehension, which is a prerequisite for generating accurate responses to human queries. We adapted an English benchmark for short-scenario language comprehension[18] into 11 additional typologically diverse languages of different sizes of speaker communities and from different language families in order to examine potential cross-linguistic variation in the comprehension abilities of three flagship LLMs (Fig.1 provides a brief overview of the research design). We aim to determine whether this variation can be attributed to some or all language-related factors previously covered in the literature: community size, type of script, and linguistic distance from English, the latter taken as the models' presumed "default language"[42] (i.e. a language that the model should use primarily and to which it may revert even when queried in other languages). Because human-model interaction relies heavily on question comprehension and answer generation, this approach lays a solid foundation for understanding how distinct languages are processed by LLMs.

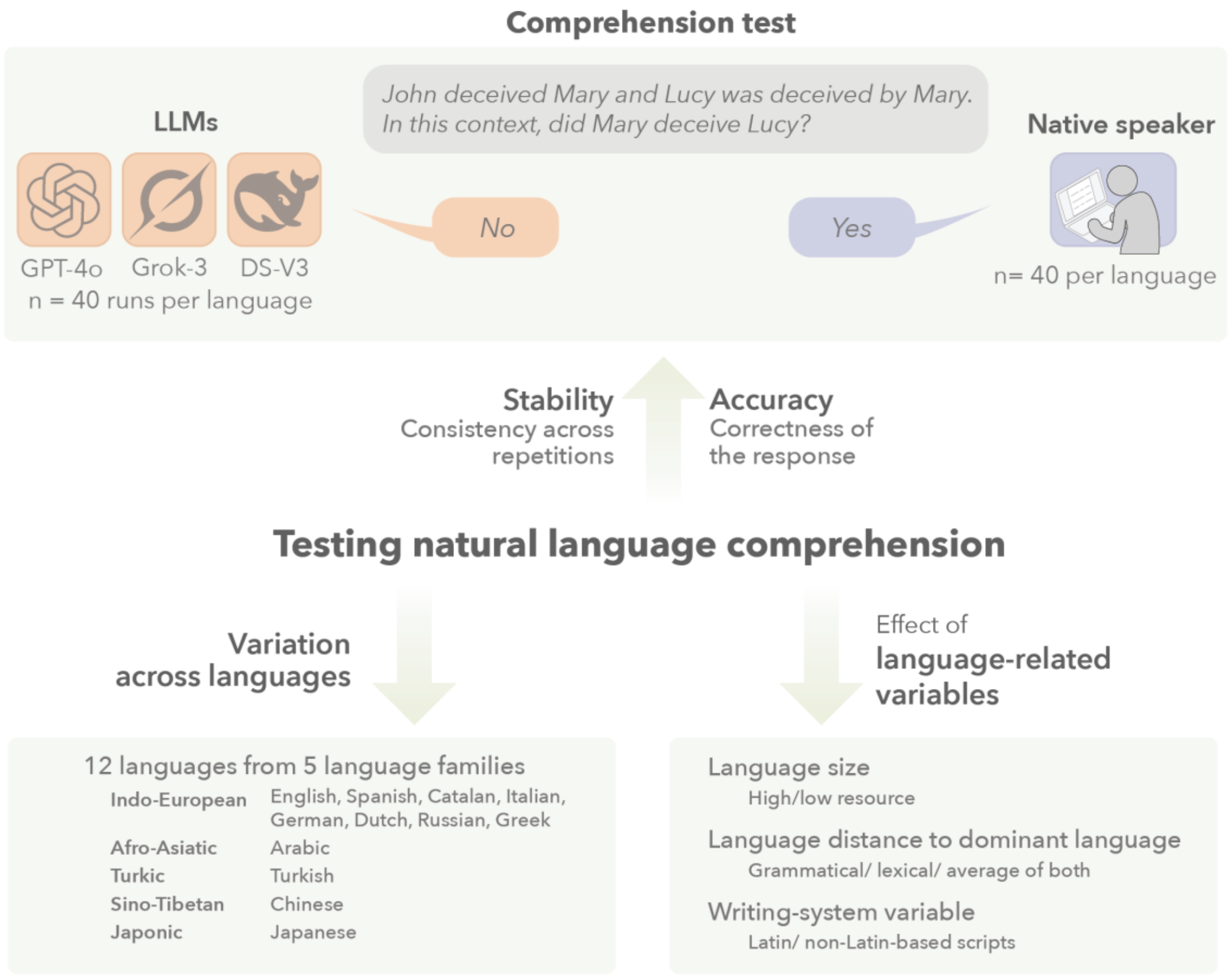

Figure 1. Research design

To this end, we address the following research questions (RQs):

RQ1. Does the language comprehension of LLMs align with human baselines across languages?

RQ2. Do the language comprehension abilities of LLMs vary across languages in terms of *accuracy* (how often the target answer is given) and *stability* (how consistent a model is in giving the target answer, when repeatedly prompted with the same question)?

RQ3. Do language size, type of script, and similarity to English explain cross-linguistic variation in LLMs' comprehension abilities?

## Results

The results are organised in two main sections, corresponding to the two primary measures of the benchmark: correctness of the response (*accuracy*) and consistency across repetitions of the same prompt (*stability*). Within each section, we first address RQ1 and RQ2 and report patterns of multilingual variation across the tested agents:

humans, GPT-4o, Grok-3, DeepSeek-V3. Subsequently, we report an analysis of potential factors that might explain this variation, tackling RQ3. Details of the experimental design and materials are provided in the Methods section, and the full dataset and the analysis code are available at https://osf.io/tvybq/overview?view_only=0d6da027a8c14aeebd9c39ed00e9970f.

**Accuracy by agent and language**

To examine response accuracy across agents and compare model performance against the human baseline, we fitted a Generalized Linear Mixed-Effects Model (GLMM) (accuracy ~ agent * lang + (1 | item) + (1 | participant_id), family = binomial). In this and all subsequent GLMMs, item denotes a unique identifier assigned to each stimulus in the benchmark, and participant_id corresponds to an individual human participant, or, for LLMs, a single model run (treated as a pseudo-participant). Model comparison using a likelihood ratio $\chi^2$ test revealed that including the interaction provided a significantly better fit for the data than the model with main effects only (accuracy ~ agent + lang + (1 | item) + (1 | participant_id); $\chi^2(33) = 1282.8$, $p < .001$), confirming the significance of the interaction and thus indicating that accuracy patterns across languages varied as a function of agent (Fig. 2).

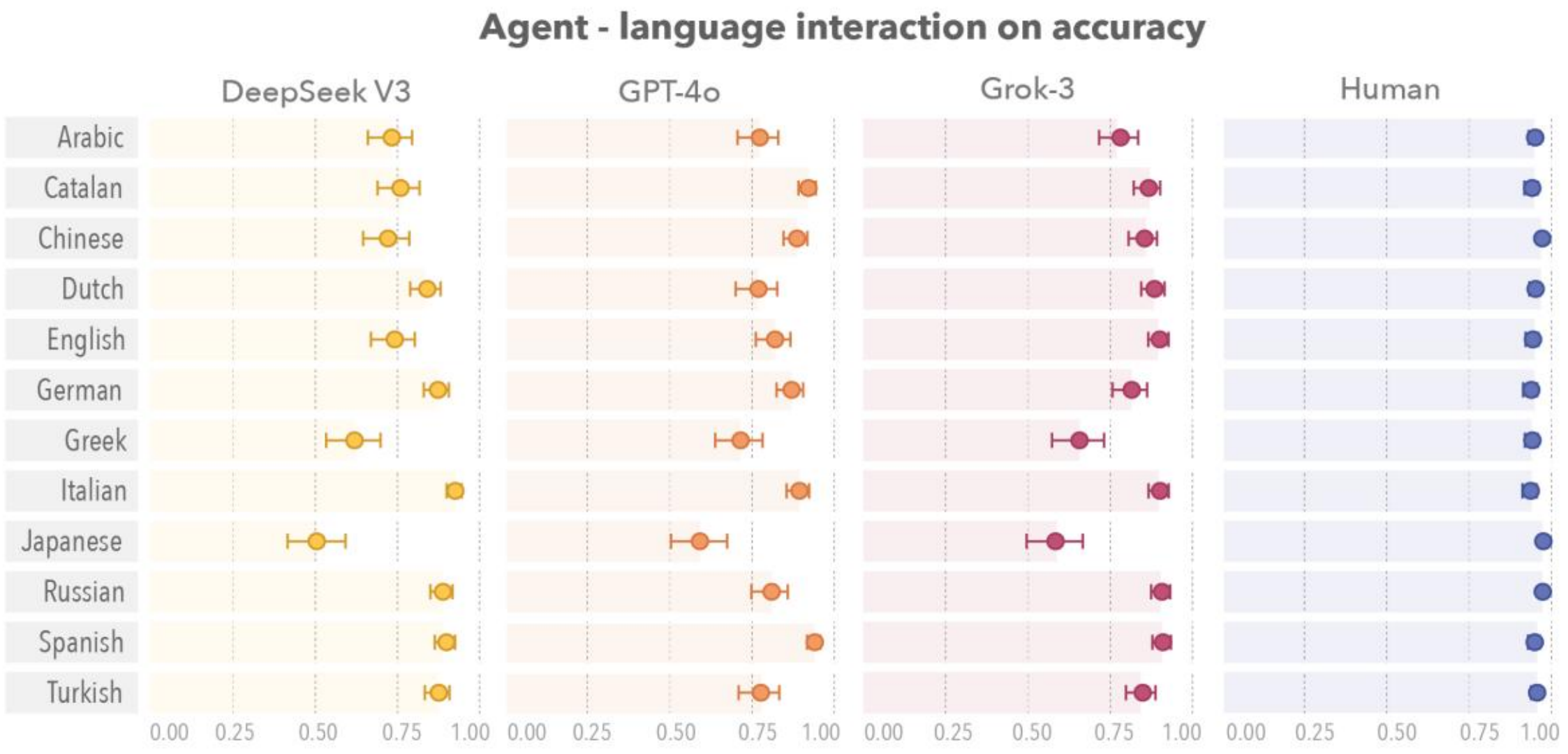

Figure 2. Accuracy across languages for each agent (estimated marginal means from GLMM with 95% confidence intervals)

To further explore this variation, post-hoc Tukey-adjusted pairwise comparisons were performed, first comparing agents within each language (~ agent | lang) and then comparing languages within each agent (~ lang | agent). Table 1 shows model-human comparisons for each language, and full statistics for all agent pairs can be found in Supplementary Table 1. Overall, humans significantly outperformed the models across languages, except for two instances: GPT-4o in Spanish (OR = 1.17, p = 0.43) and DeepSeek-V3 in Italian (OR = 1.07, p = 0.904), where the models' accuracy was comparable to that of human participants.

| Lang | Agent 1 | Agent 2 | Odds | SE | z | p |
|---|---|---|---|---|---|---|
| **Arabic** | Humans | DS_V3 | 6.90 | 0.57 | 23.29 | <.001 |
| | | GPT-4o | 5.71 | 0.47 | 21.17 | <.001 |
| | | Grok-3 | 5.31 | 0.44 | 20.02 | <.001 |
| **Catalan** | Humans | DS_V3 | 4.97 | 0.46 | 17.3 | <.001 |
| | | GPT-4o | 1.36 | 0.13 | 3.13 | 0.009 |
| | | Grok-3 | 2.37 | 0.22 | 9.12 | <.001 |
| **Chinese** | Humans | DS_V3 | 13.46 | 1.40 | 25.06 | <.001 |
| | | GPT-4o | 4.54 | 0.48 | 14.17 | <.001 |
| | | Grok-3 | 5.85 | 0.62 | 16.74 | <.001 |
| **Dutch** | Humans | DS_V3 | 3.61 | 0.35 | 13.22 | <.001 |
| | | GPT-4o | 6.01 | 0.57 | 18.73 | <.001 |
| | | Grok-3 | 2.5 | 0.24 | 9.14 | <.001 |
| **English** | Humans | DS_V3 | 5.60 | 0.51 | 18.75 | <.001 |
| | | GPT-4o | 3.64 | 0.34 | 13.83 | <.001 |
| | | Grok-3 | 1.72 | 0.17 | 5.59 | <.001 |
| **German** | Humans | DS_V3 | 2.10 | 0.20 | 7.80 | <.001 |
| | | GPT-4o | 2.29 | 0.22 | 8.70 | <.001 |
| | | Grok-3 | 3.38 | 0.32 | 12.96 | <.001 |
| **Greek** | Humans | DS_V3 | 9.85 | 0.91 | 24.73 | <.001 |
| | | GPT-4o | 6.54 | 0.60 | 20.27 | <.001 |
| | | Grok-3 | 8.42 | 0.78 | 23.02 | <.001 |
| **Italian** | Humans | DS_V3 | 1.07 | 0.10 | 0.68 | 0.904 |
| | | GPT-4o | 1.75 | 0.17 | 5.89 | <.001 |
| | | Grok-3 | 1.53 | 0.15 | 4.43 | .0001 |
| **Japanese** | Humans | DS_V3 | 39.49 | 4.20 | 34.52 | <.001 |
| | | GPT-4o | 28.43 | 3.02 | 31.47 | <.001 |
| | | Grok-3 | 28.79 | 3.06 | 31.59 | <.001 |
| **Russian** | Humans | DS_V3 | 4.42 | 0.48 | 13.77 | <.001 |
| | | GPT-4o | 8.93 | 0.94 | 20.74 | <.001 |
| | | Grok-3 | 3.61 | 0.39 | 11.80 | <.001 |
| **Spanish** | Humans | DS_V3 | 1.96 | 0.19 | 6.82 | <.001 |
| | | GPT-4o | 1.17 | 0.12 | 1.50 | 0.435 |
| | | Grok-3 | 1.73 | 0.17 | 5.49 | <.001 |
| **Turkish** | Humans | DS_V3 | 2.97 | 0.29 | 11.03 | <.001 |
| | | GPT-4o | 6.42 | 0.62 | 19.31 | <.001 |
| | | Grok-3 | 3.82 | 0.37 | 13.65 | <.001 |

Table 1: Human-LLMs comparisons per language

Additionally, GPT-4o showed the highest overall accuracy, outperforming the other models in four of the tested languages (Spanish, Catalan, Chinese, and Greek) and sharing the leading position with another model in three more (Arabic and Japanese with Grok-3, German with DeepSeek-V3). Grok-3 held the lead in two languages (English and Dutch) and tied in three (Russian with DeepSeek-V3, Arabic and Japanese with GPT-4o), whereas DeepSeek-V3 led in two (Italian and Turkish) and tied in two (German with GPT-4o and Russian with Grok-3).

**Cross-linguistic variation for each agent**

Table 2 reports estimated marginal means and standard errors for each tested language and model. Pairwise comparisons revealed a greater number of significant cross-linguistic differences in LLMs than in humans. Although some contrasts between languages did emerge as significant in the human data, these differences were small and mainly involved four languages —Japanese, Chinese, Russian, and Turkish— that exhibited lower individual variation than the rest (standard errors of 0.005, 0.005, 0.005, 0.008, respectively). Japanese, Chinese, and Russian had slightly higher accuracy than the rest of the languages (but were comparable between themselves), and Turkish was marginally better than Italian and German (full statistics for pairwise comparisons of languages within each agent can be found in Supplementary Table 2).

| | **Agent** | | | |
|---|---|---|---|---|
| **Language** | **Humans** | **GPT-4o** | **Grok-3** | **DS_V3** |
| **Arabic** | 0.95 (0.009) | 0.77 (0.032) | 0.79 (0.030) | 0.74 (0.035) |
| **Catalan** | 0.94 (0.01) | 0.92 (0.013) | 0.87 (0.02) | 0.77 (0.033) |
| **Chinese** | 0.97 (0.005) | 0.89 (0.018) | 0.86 (0.022) | 0.73 (0.036) |
| **Dutch** | 0.95 (0.009) | 0.77 (0.032) | 0.89 (0.018) | 0.85 (0.024) |
| **English** | 0.94 (0.01) | 0.82 (0.027) | 0.91 (0.016) | 0.75 (0.034) |
| **German** | 0.94 (0.011) | 0.87 (0.021) | 0.82 (0.027) | 0.88 (0.019) |
| **Greek** | 0.94 (0.01) | 0.72 (0.037) | 0.66 (0.041) | 0.63 (0.043) |
| **Italian** | 0.94 (0.011) | 0.89 (0.017) | 0.91 (0.016) | 0.93 (0.012) |
| **Japanese** | 0.98 (0.005) | 0.59 (0.044) | 0.59 (0.044) | 0.51 (0.046) |
| **Russian** | 0.97 (0.005) | 0.81 (0.028) | 0.91 (0.015) | 0.90 (0.017) |
| **Spanish** | 0.95 (0.009) | 0.94 (0.01) | 0.92 (0.014) | 0.91 (0.016) |
| **Turkish** | 0.96 (0.008) | 0.78 (0.032) | 0.85 (0.023) | 0.88 (0.019) |

Table 2: Estimated marginal means and standard error (in brackets) for tested languages across models

In contrast to humans, the performance of LLMs varied more pronouncedly across languages. Crucially, English was not the best-performing language for any of the tested models. DeepSeek-V3 reached the highest accuracy in Italian, followed by Spanish, Russian, and Turkish. GPT-4o performed best in Spanish and Catalan, followed by Italian and Chinese. In both cases, English ranked mid-range ($8^{th}$ for DeepSeek-V3, $6^{th}$ for GPT-4o). Grok-3 demonstrated the least amount of cross-linguistic variation, with no significant differences among its five top-performing languages (i.e. Spanish, Italian, Russian, English, and Dutch). Even though overall cross-linguistic patterns were not identical for the tested LLMs, certain similarities could be observed: Greek and Japanese consistently showed the lowest accuracy across the models, while Spanish and Italian appeared among the best-performing languages.

To identify potential driving forces of cross-linguistic variation in LLM performance, we modelled accuracy as a function of three language-related factors: language size, writing system, and language distance from the expected and actual best-performing languages (English and Spanish, respectively, as established above). To this end, a GLMM (accuracy ~ distance_spanish + distance_english + writing_system*lang_size + (1 | item), family = binomial) was fitted for each LLM. The interaction between writing system and language size was included based on preliminary exploratory analyses.

The optimal type of language-distance measure (lexical, grammatical, or average of the two) was first determined for each LLM separately through systematic model comparisons based on Akaike Information Criteria (AIC)[43], which balances model fit and complexity to determine optimal models. Since lexical distances offered a slightly better fit in most cases, they were selected for the final predictive models.

For GPT-4o, the fitted statistical model returned a significant negative main effect of language distance from Spanish ($\beta = -0.6$, SE = 0.02, $z = -36.2$, $p < .001$), while the effect of distance from English was significant and positive ($\beta = 0.096$, SE = 0.02, $z = 5.8$, $p < .001$), suggesting that languages closer to Spanish demonstrated higher accuracy, whereas accuracy increased as distance from English grew. Additionally, language size had a significantly stronger impact on accuracy for languages with non-Latin-based scripts ($\beta = 0.6$, SE = 0.03, $z = 21.6$, $p < .001$), than for Latin-based alphabets (simple effect: $\beta = 0.05$, SE = 0.017, $z = 3.28$, $p = 0.001$).

For Grok, distance from Spanish emerged as a significant negative predictor for accuracy ($\beta = -0.4$, SE = 0.02, $z = -23.8$, $p < .001$), as well as distance from English, although with a smaller magnitude ($\beta = -0.16$, SE = 0.02, $z = -8.6$, $p < .001$). The interaction between writing system and language size followed a pattern similar to that of GPT-4o, with more pronounced effects of language size for non-Latin-based scripts ($\beta = 0.6$, SE = 0.03, $z = 18.7$, $p < .001$), compared to Latin-based writing systems (simple effect: $\beta = 0.08$, SE = 0.02, $z = 4.12$, $p < .001$).

The results of DeepSeek-V3 showed a similar trend: smaller distances from Spanish ($\beta = -0.38$, SE = 0.02, $z = -23.9$, $p < .001$) and English ($\beta = -0.07$, SE = 0.02, $z = -4.2$, $p < .001$) were associated with better performance, with the effect being stronger for Spanish. Consistent with the other models, the slope of language size was significantly steeper for non-Latin-based scripts ($\beta = 0.5$, SE = 0.03, $z = 20.2$, $p < .001$), compared to Latin-based ones (simple effect: $\beta = -0.25$, SE = 0.017, $z = -14.79$, $p < .001$).

Overall, the accuracy analyses reveal three consistent trends across models: (i) languages more similar to Spanish performed better across all the tested LLMs; (ii) similarity to English either had a weaker or opposite effect on accuracy; and (iii) the effect of language size varied depending on the script, with a more pronounced impact for non-Latin writing systems.

**Stability by agent and language**

The analyses for stability across languages and agents followed the same structure as the one reported above for accuracy. First, a GLMM was built (stability ~ agent * lang + (1 | item) + (1 | participant_id), family = binomial). The interaction model was significantly better than the model with the main effects only (stability ~ agent + lang + (1 | item) + (1 | participant_id); $\chi^2(33) = 639.73$, $p < .001$), indicating the significance of the interaction term and confirming substantial cross-linguistic and cross-agent differences. The overall interaction pattern is shown in Fig. 3.

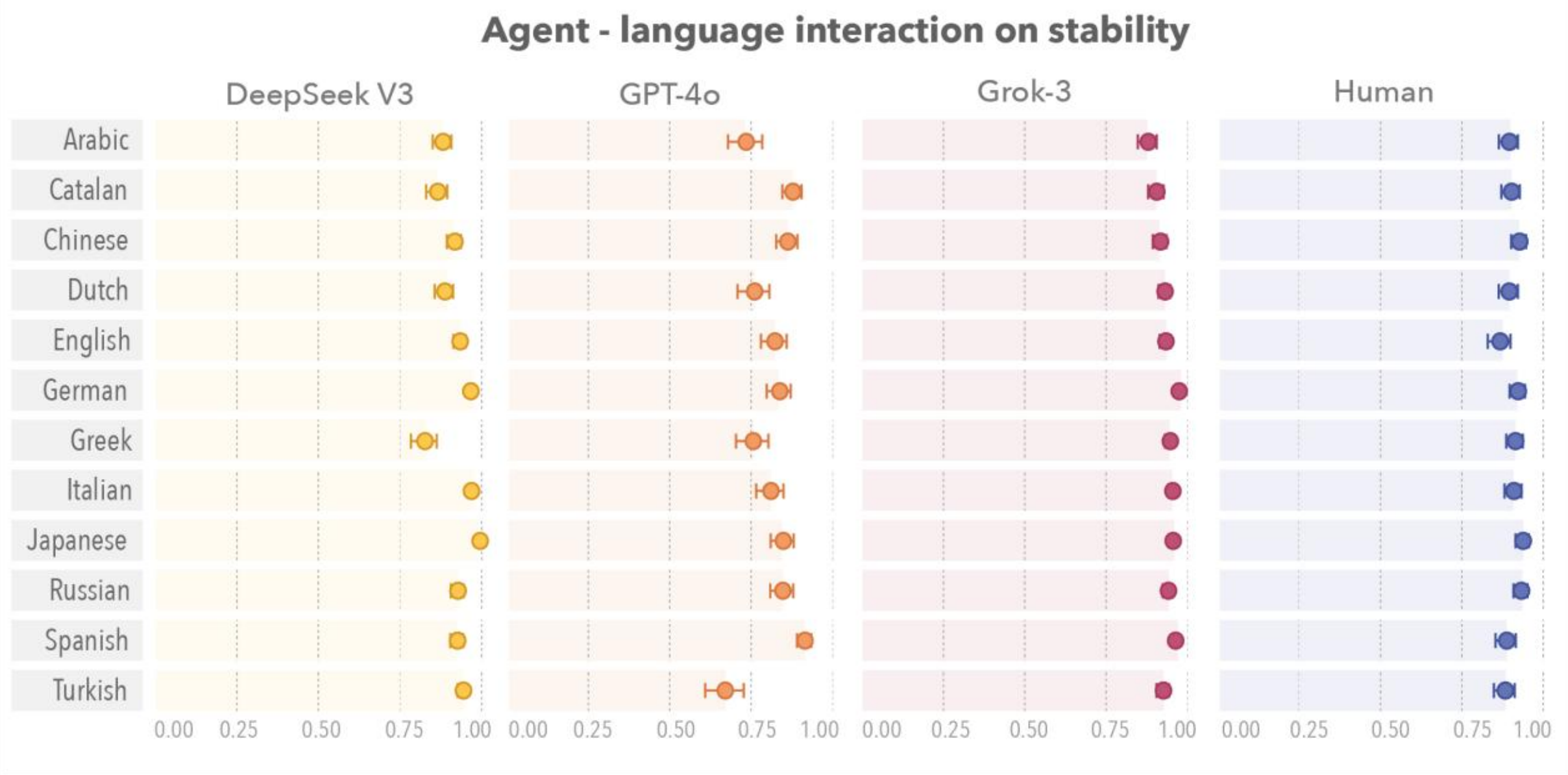

Figure 3. Stability across languages for each agent (estimated marginal means from GLMM with 95% confidence intervals)

Tukey-adjusted pairwise comparisons were then computed to further explore this variation (full statistics for pairwise contrasts can be found in Supplementary Tables 3 and 4).

The analysis showed that humans outperform GPT-4o in terms of providing stable responses in most languages, with the exception of English and Catalan, where the two agents showed comparable performance, and Spanish, where GPT-4o demonstrated higher stability than humans. However, DeepSeek-V3 was significantly more stable in its responses than human participants in 7 out of 12 tested languages and performed comparably in 4 of them (Arabic, Catalan, Chinese, and Russian), scoring significantly below the human group only in Greek. Grok-3 exhibited even higher stability, outperforming humans in 9 languages and showing no significant difference from them in Catalan, Chinese, and Arabic. Thus, performance of both Grok-3 and DeepSeek-V3 was overall more stable than the human baseline, with Grok-3 showing the highest stability level across all the agents. Only GPT-4o appeared to be consistently inferior to humans.

Moreover, Grok-3 demonstrated the least amount of cross-linguistic variation in stability, followed by DeepSeek-V3, while considerable differences across languages were observed for GPT-4o. Unlike accuracy, stability in human responses was less homogeneous cross-linguistically.

**Explaining variability in stability**

The role of language-related factors in stability variation across languages was explored in a similar fashion to the accuracy analysis.

For DeepSeek-V3, both distance from Spanish ($\beta = 0.17$, SE = 0.05, z = 3.5, $p < .001$) and distance from English ($\beta = 0.15$, SE = 0.05, z = 2.8, $p < .001$) had significant positive effects, indicating decreased stability for languages closer to these two reference points. Additionally, no significant interaction between language size and type of writing system was observed for this model ($\beta = 0.15$, SE = 0.09, z = -1.7.8, p = 0.08), and main effect of language size was positive and significant ($\beta = 0.129$, SE = 0.040, z = 3.25, p = 0.001).

For Grok-3, a significant negative effect of distance from Spanish ($\beta = –0.77$, SE= 0.04, z= -1.98, p = 0.048 and English ($\beta = –0.071$, SE= 0.05, z= -1.52, p = 0.129) was observed. A significant interaction ($\beta = –0.14$, SE= 0.07, z= -2.1, p = 0.033) between language size and writing system type revealed a negative effect of language size on stability for the scripts not based on the Latin alphabet, while it had no significant effect for Latin-based scripts ($\beta = –0.03$, SE = 0.046, z = –0.57, p = 0.57).

For GPT-4o, distance from Spanish had a strong negative impact on stability ($\beta = –0.38$, SE= 0.03, z= -14.9, $p < .001$), while distance from English was not significant ($\beta = –0.05$, SE = 0.03, z = –2.2, p = 0.07). Regarding the interaction with writing system, language size had a significantly stronger positive effect for non-Latin-based scripts ($\beta = 0.19$, SE = 0.05, z = 4.04, $p < .001$) compared to the reference system (Latin alphabet, simple effect: $\beta = 0.098$, SE = 0.028, z = 3.48, $p < .001$).

## Discussion

The present study set out to test the cross-linguistic performance of LLMs on the aspect of language that lies at the very core of their functionality: the ability to comprehend and provide answers to written prompts. The overall aim is to determine how uniformly LLMs from different families perform across languages and what factors potentially drive any possible variation. As much research evokes the black-box nature of LLMs[44], understanding what drives their performance and how stable this performance eventually is across different languages is of paramount importance. To this end, our experiment was guided by three main RQs:

RQ1. Does LLMs' language comprehension align with human baselines across languages?

RQ2. Do LLMs' comprehension abilities vary across languages in terms of *accuracy* (how often the target answer is given) and *stability* (how consistent a model is in giving the target answer, when repeatedly prompted with the same question)?

RQ3. Do language size, type of script, and similarity to English explain cross-linguistic variation in LLMs' comprehension abilities?

To address these questions, we extended an existing benchmark for English comprehension[18] to 11 additional languages and evaluated the performance of three flagship models against human baselines, focusing on the accuracy and stability of their responses.

Starting with RQ1, and in line with previous findings[18], our results on accuracy show that even top-performing models fail to reach human-like levels of comprehension across a variety of languages, highlighting the potential risks of over-reliance on LLM-powered technology. Out of 12 tested languages and three models, LLMs approximated human accuracy in only two cases: Spanish for GPT-4o and Italian for DeepSeek-V3. This is perhaps unexpected given the nature of the task: answering questions based on a user-provided prompt is central to the models' intended use and therefore amounts to a core competence they could be reasonably expected to exhibit consistently across languages. If LLMs struggle with general language comprehension tasks even in high-resource Indo-European languages, this might signal deeper structural limitations in their linguistic representations. Our results on accuracy are therefore consistent with prior work documenting significant limitations in the current language abilities of LLMs.[14-18]

Regarding RQ2, our results also align with previous reports of cross-linguistic disparities in model performance,[30-32] revealing significant variation in LLMs' language comprehension abilities. However, our results diverge from the widespread assumption —common in previous studies[33-37]— that English constitutes the dominant or default language of LLMs. Specifically, contrary to Zhang et al.'s[37] interpretation of a subordinate multilingual organisation in LLMs, with English as the representational core, this language was not the strongest performer in any of the models tested. Its best performance was observed in Grok-3, where it appeared among the five best-performing languages alongside Spanish, Russian, Italian, and Dutch, but it ranked only sixth and eighth in GPT-4o and DeepSeek-V3 respectively. In contrast, Spanish and Italian consistently scored at or near ceiling for all tested LLMs. These findings partially echo Kim et al.[41], who found that Romance and Slavic languages outperformed English in information retrieval and aggregation tasks, pointing against English being the default language of the models.

In response to RQ3, a closer examination of the observed cross-linguistic patterns and their underlying factors further revealed that languages more similar to Spanish tended to perform better, whereas similarity to English had a small or even a negative effect in terms of accuracy. While additional data would be needed to draw definitive conclusions, this asymmetry suggests that different language systems may entail varying degrees of difficulty for LLMs. One possible explanation, also touched upon in Kim et al.[41], is tokenization. While the amount of training data alone cannot account for the fact that Spanish and Italian consistently outperformed English and German, it is plausible that current tokenization mechanisms are more linguistically efficient for some languages. Supporting this possibility, Kim et al.[41] showed that when controlling for the amount of information received by the models rather than the number of tokens, the relative ranking of languages changed, with English performance declining further and Slavic languages outperforming the Romance group. This suggests that tokenization too, and not raw data size alone, may contribute to cross-linguistic performance differences.

Our results also showed that the impact of language size interacts with the type of writing system: differences between high- and low-resource languages were more pronounced in non-Latin-based scripts than in Latin-based ones. This indicates that languages with non-Latin-based writing systems require substantially larger datasets for models to achieve comparable levels of accuracy. The disadvantage of non-Latin scripts is further confirmed by the consistently low scores for Greek and Japanese across models, likely reflecting the combined effects of script type and limited data. Since many languages spoken by non-WEIRD populations employ non-Latin scripts and have comparatively small speaker communities, this limitation can have important implications for the capabilities of LLMs and raises concerns about equitable access to reliable AI tools.

The multilingual patterns observed for the second benchmark dimension (i.e. stability) differed significantly from those found for accuracy, suggesting that although they were brought together in the original metric, they reflect different aspects of models' performance, with temperature setting being one of them. Unlike accuracy, two out of the three models outperformed human participants in terms of stability, and considerable variation across LLMs could be observed. This indicates that while human performance can be affected by attentional lapses, fatigue, or cognitive load, LLMs —when operating under certain temperature conditions— can maintain highly consistent outputs throughout linguistically demanding tasks.

Taken together, our findings show that LLM outcomes vary considerably depending on the prompting language even for a simple comprehension task, potentially creating disparities in information representation across linguistic communities, with non-WEIRD populations being at greater risk for receiving compromised or unclear information. In particular, several non-WEIRD languages (e.g., Greek, Japanese, and, to a certain extent, Arabic) were linked to compromised linguistic comprehension abilities in LLMs; a finding linked to script systems and data availability. Crucially, however, our results reveal that WEIRDness does not guarantee accuracy or reliability of linguistic information: English and German —prototypical WEIRD, high-resource languages— were not the top-performing languages. In this sense, we discover a Romance puzzle in the linguistic abilities of AI models: Romance languages, such as Spanish and Italian, exhibit higher performance than assumed top-performers, namely English and other languages from the Germanic group. This asymmetry implies that even speakers of high-resource languages, that are predominantly used in WEIRD communities, may face elevated risks of misinterpretation or misinformation; a finding that invites closer examination of the linguistic capabilities of LLMs. More broadly, our results demonstrate that LLMs still fall short of human-like comprehension even in languages with extensive training data.

While these findings provide valuable insights into the linguistic capabilities of LLMs, several methodological limitations should be acknowledged. First, language comprehension was tested on a limited set of linguistic structures. Expanding this set might offer a more complete picture of how models interpret human language. Similarly, a more balanced testing across language families, sizes, and writing systems might help determine the relative influence of each factor. A different operationalisation of these factors may also yield a more nuanced understanding of their effects. Finally, only three closed models were tested in this experiment. Future work could investigate open models run locally or trained from scratch to better isolate architectural and training-related effects.

# Methods

To collect English data, the language comprehension benchmark of Dentella et al. (2024) was used.[18] It includes 40 target items and 2 attention checks, with each item consisting of a short scenario (1-2 sentences), followed by a Yes-No comprehension question. All sentences are affirmative, contain high-frequency vocabulary and follow a coordination pattern, as exemplified in (1).

(1) John deceived Mary and Lucy was deceived by Mary. In this context, did Mary deceive Lucy?

Proper names are used instead of pronouns to prevent pronoun resolution from interfering with the comprehension task. This ensures sufficient processing difficulty without introducing unnecessary grammatical complexity.

For the purposes of the present study, the original English dataset was adapted into 11 additional languages from a variety of language families: Spanish, Catalan, Italian, German, Dutch, Russian, Greek (Indo-European), Arabic (Afro-Asiatic), Turkish (Turkic), Chinese (Sino-Tibetan), and Japanese (Japonic). The selected languages were balanced in terms of WEIRDness (6 languages predominantly associated with WEIRD and 6 languages predominantly associated with non-WEIRD communities). Moreover, the tested languages differed in the size of language community as well as the type of writing system, creating a diverse range of language systems for a comprehensive comparison.

The adaptations of the original benchmark across languages were implemented by trained linguists who were native speakers of the respective languages. These were instructed to keep the translation as close as possible to the original, while also preserving grammaticality and naturalness. Since direct translations were not always feasible due to language-specific constraints (e.g., intransitive verbs cannot be used in the passive voice in Russian), lexical adjustments were made in such cases (e.g., “was helped” was translated as “was saved”). The changes were limited to the use of synonyms, semantically related words or high-frequency alternatives. Lexical adjustments were preferred over structural ones because they were expected to have minimal impact on processing for both humans and LLMs. Translators were also asked to consult each other’s adaptation and harmonise choices across languages whenever possible, in order to limit unnecessary cross-linguistic variation.

## LLM data

Three flagship models were tested: GPT-4o[45], Grok-3[46], and DeepSeek-V3[47]. GPT-4o was chosen as the most advanced representative (at the time of testing) of the GPT family, which previously achieved the strongest performance on the English benchmark[18]. Grok-3 and DeepSeek-V3 were included as additional state-of-the-art language models developed by prominent industry groups that have not been previously tested on the benchmark.

The models were prompted via their respective APIs using the OpenAI SDK.[48] The Python code used for prompting is available at https://osf.io/tvybq/overview?view_only=0d6da027a8c14aeebd9c39ed00e9970f. To avoid data imbalance in human-LLM comparison, each model was prompted with the same data set through 40 independent runs, yielding 40 pseudo-participants per model. That resulted in the same number of data points per language for both agents, LLMs and human (5040; see the Procedure section below for more details). Default temperature setting was used for all the models to approximate typical user experience offered by the interface and to avoid discrepancies across the tested models, since they use different scales for the given parameter.

**Human data**

For the human baseline, we collected data from 480 native speakers (n= 40 per language) recruited through Prolific. A full summary of the demographic data is available at https://osf.io/tvybq/overview?view_only=0d6da027a8c14aeebd9c39ed00e9970f. 39 participants were excluded during the initial data screening due to incomplete responses, failed attention checks, or atypical reaction times (shorter than one second or longer than one minute). Replacement participants were subsequently recruited to maintain balanced sample sizes.

To complete the study, participants submitted their responses in a self-paced online task by typing into a text field displayed beneath each question. The task was designed using the jsPsych framework[49] (jsPsych v8) and hosted on the MindProbe server. A consent form was presented at the beginning of the experiment and participants were paid fairly in accordance with Prolific guidelines. All procedures adhered to the Declaration of Helsinki and were reviewed and approved by the Research Ethics Committee at the authors' home institution.

**Procedure**

Since stability of models' responses over the repetition of the same prompt was part of the evaluation metric used in the study, each test item was repeated 3 times, resulting in the final dataset of 126 trials per language. All the trials were presented one at a time in a randomised order for humans as well as for LLMs. The instruction "answer using just one word" was added after the question in each trial, since the one-word setting was shown to be more favourable for LLMs than the open-length format for the original benchmark[18] and was thus considered to offer optimal conditions for a fair evaluation of models' capabilities.

**Scoring**

Accuracy was scored for each trial: A value of "1" was assigned to correct answers, and "0" to incorrect ones. Minor variability (e.g., typos, "Mary", "no one", "yep", etc.) was not penalised as long as the answer could be unambiguously interpreted as "yes" or "no".

All uncertain responses (e.g., “unclear”, “impossible to answer”, “maybe”) were scored as incorrect.

Stability was calculated for each item and participant, with a value of “1” assigned when all three trials from a given item and participant had identical accuracy scores (either 1 or 0), and “0” when at least one trial differed. The dataset was subsequently aggregated to retain one stability value for each item–participant combination.

**Analysis**

All statistical analyses were performed in R version 4.4.3[50], and the following packages were used: lme4[51] and emmeans[52].

**Language-related variables**

Language size was operationalised as the total number of speakers, based on the data from Ethnologue.[53] Language distance was used as the measure of similarities between languages, with lower distance values corresponding to a higher number of shared features. Three types of distances were included in the analysis: grammatical, lexical, and the average of the two. Grammatical distances were extracted from Grambank[54] and lexical ones from eLinguistics.[55] The writing-system variable included the following levels: Latin-based and non-Latin-based (that covered other alphabetic, abjad, and logographic scripts). To ensure comparability of coefficients and prevent variables measured on different scales from dominating the analyses, all language-related factors were standardised using R’s scale() function.

## Supplementary Information

### Supplementary Table 1: Pairwise comparisons of all agents per language (accuracy)

| Language | Agent 1 | Agent 2 | OR | SE | z | p |
|---|---|---|---|---|---|---|
| **Arabic** | human | GPT-4o | 5.706 | 0.469 | 21.17 | <.001 |
| | human | Grok-3 | 5.307 | 0.442 | 20.03 | <.001 |
| | human | DS_V3 | 6.902 | 0.573 | 23.29 | <.001 |
| | DS_V3 | GPT-4o | 0.827 | 0.067 | -2.35 | 0.088 |
| | DS_V3 | Grok-3 | 0.769 | 0.063 | -3.22 | 0.007 |
| | GPT-4o | Grok-3 | 0.93 | 0.076 | -0.89 | 0.810 |
| **Catalan** | human | GPT-4o | 1.36 | 0.134 | 3.13 | 0.010 |
| | human | Grok-3 | 2.375 | 0.225 | 9.12 | <.001 |
| | human | DS_V3 | 4.979 | 0.462 | 17.3 | <.001 |
| | DS_V3 | GPT-4o | 0.273 | 0.025 | -14.29 | <.001 |
| | DS_V3 | Grok-3 | 0.477 | 0.042 | -8.49 | <.001 |
| | GPT-4o | Grok-3 | 1.746 | 0.163 | 5.99 | <.001 |
| **Chinese** | human | GPT-4o | 4.539 | 0.485 | 14.17 | <.001 |
| | human | Grok-3 | 5.847 | 0.617 | 16.74 | <.001 |
| | human | DS_V3 | 13.462 | 1.397 | 25.06 | <.001 |
| | DS_V3 | GPT-4o | 0.337 | 0.03 | -12.41 | <.001 |
| | DS_V3 | Grok-3 | 0.434 | 0.038 | -9.66 | <.001 |
| | GPT-4o | Grok-3 | 1.288 | 0.116 | 2.81 | 0.025 |
| **Dutch** | human | GPT-4o | 6.008 | 0.575 | 18.73 | <.001 |
| | human | Grok-3 | 2.475 | 0.245 | 9.14 | <.001 |
| | human | DS_V3 | 3.611 | 0.351 | 13.22 | <.001 |
| | DS_V3 | GPT-4o | 1.664 | 0.144 | 5.9 | <.001 |
| | DS_V3 | Grok-3 | 0.685 | 0.062 | -4.2 | <.001 |
| | GPT-4o | Grok-3 | 0.412 | 0.036 | -10.04 | <.001 |
| **English** | human | GPT-4o | 3.639 | 0.34 | 13.83 | <.001 |
| | human | Grok-3 | 1.718 | 0.166 | 5.59 | <.001 |
| | human | DS_V3 | 5.605 | 0.515 | 18.75 | <.001 |
| | DS_V3 | GPT-4o | 0.649 | 0.055 | -5.07 | <.001 |
| | DS_V3 | Grok-3 | 0.306 | 0.027 | -13.29 | <.001 |
| | GPT-4o | Grok-3 | 0.472 | 0.043 | -8.32 | <.001 |
| **German** | human | GPT-4o | 2.286 | 0.217 | 8.7 | <.001 |
| | human | Grok-3 | 3.378 | 0.317 | 12.97 | <.001 |
| | human | DS_V3 | 2.097 | 0.199 | 7.79 | <.001 |
| | DS_V3 | GPT-4o | 1.09 | 0.098 | 0.96 | 0.774 |
| | DS_V3 | Grok-3 | 1.611 | 0.143 | 5.38 | <.001 |
| | GPT-4o | Grok-3 | 1.478 | 0.13 | 4.42 | <.001 |
| **Greek** | human | GPT-4o | 6.539 | 0.606 | 20.27 | <.001 |
| | human | Grok-3 | 8.423 | 0.78 | 23.02 | <.001 |
| | human | DS_V3 | 9.855 | 0.912 | 24.73 | <.001 |
| | DS_V3 | GPT-4o | 0.664 | 0.055 | -4.93 | <.001 |
| | DS_V3 | Grok-3 | 0.855 | 0.071 | -1.9 | 0.230 |
| | GPT-4o | Grok-3 | 1.288 | 0.107 | 3.04 | 0.013 |
| **Italian** | human | GPT-4o | 1.755 | 0.167 | 5.89 | <.001 |
| | human | Grok-3 | 1.532 | 0.147 | 4.43 | <.001 |
| | human | DS_V3 | 1.07 | 0.106 | 0.68 | 0.904 |
| | DS_V3 | GPT-4o | 1.64 | 0.157 | 5.17 | <.001 |
| | DS_V3 | Grok-3 | 1.432 | 0.138 | 3.72 | 0.001 |
| | GPT-4o | Grok-3 | 0.873 | 0.081 | -1.46 | 0.460 |
| **Japanese** | human | GPT-4o | 28.428 | 3.023 | 31.47 | <.001 |
| | human | Grok-3 | 28.789 | 3.062 | 31.59 | <.001 |
| | human | DS_V3 | 39.489 | 4.205 | 34.52 | <.001 |
| | DS_V3 | GPT-4o | 0.72 | 0.059 | -3.99 | <.001 |
| | DS_V3 | Grok-3 | 0.729 | 0.06 | -3.83 | <.001 |
| | GPT-4o | Grok-3 | 1.013 | 0.084 | 0.15 | 0.999 |
| **Russian** | human | GPT-4o | 8.933 | 0.943 | 20.74 | <.001 |
| | human | Grok-3 | 3.608 | 0.392 | 11.81 | <.001 |
| | human | DS_V3 | 4.423 | 0.478 | 13.77 | <.001 |
| | DS_V3 | GPT-4o | 2.019 | 0.18 | 7.86 | <.001 |
| | DS_V3 | Grok-3 | 0.816 | 0.076 | -2.18 | 0.129 |
| | GPT-4o | Grok-3 | 0.404 | 0.037 | -10.01 | <.001 |
| **Spanish** | human | GPT-4o | 1.166 | 0.119 | 1.5 | 0.435 |
| | human | Grok-3 | 1.728 | 0.172 | 5.49 | <.001 |
| | human | DS_V3 | 1.958 | 0.193 | 6.83 | <.001 |
| | DS_V3 | GPT-4o | 0.596 | 0.058 | -5.29 | <.001 |
| | DS_V3 | Grok-3 | 0.883 | 0.084 | -1.32 | 0.551 |
| | GPT-4o | Grok-3 | 1.482 | 0.147 | 3.98 | <.001 |
| **Turkish** | human | GPT-4o | 6.417 | 0.618 | 19.31 | <.001 |
| | human | Grok-3 | 3.818 | 0.375 | 13.65 | <.001 |
| | human | DS_V3 | 2.967 | 0.293 | 11.03 | <.001 |
| | DS_V3 | GPT-4o | 2.162 | 0.19 | 8.78 | <.001 |
| | DS_V3 | Grok-3 | 1.287 | 0.115 | 2.81 | 0.025 |
| | GPT-4o | Grok-3 | 0.595 | 0.052 | -5.98 | <.001 |

**Supplementary Table 2: Pairwise comparisons of languages within each agent (accuracy)**

| Agent | Lang1 | Lang2 | OR | SE | z | p |
|---|---|---|---|---|---|---|
| human | Arabic | Catalan | 1.2 | 0.113 | 1.94 | 0.736 |
| human | Arabic | Chinese | 0.544 | 0.057 | -5.81 | <.001 |
| human | Arabic | Dutch | 0.973 | 0.095 | -0.28 | 1.000 |
| human | Arabic | English | 1.173 | 0.11 | 1.7 | 0.870 |
| human | Arabic | German | 1.264 | 0.12 | 2.47 | 0.362 |
| human | Arabic | Greek | 1.19 | 0.114 | 1.82 | 0.807 |
| human | Arabic | Italian | 1.302 | 0.121 | 2.85 | 0.161 |
| human | Arabic | Japanese | 0.477 | 0.052 | -6.82 | <.001 |
| human | Arabic | Russian | 0.513 | 0.054 | -6.32 | <.001 |
| human | Arabic | Spanish | 1.031 | 0.099 | 0.32 | 1.000 |
| human | Arabic | Turkish | 0.873 | 0.086 | -1.38 | 0.967 |
| human | Catalan | Chinese | 0.453 | 0.05 | -7.15 | <.001 |
| human | Catalan | Dutch | 0.81 | 0.084 | -2.04 | 0.666 |
| human | Catalan | English | 0.977 | 0.098 | -0.23 | 1.000 |
| human | Catalan | German | 1.053 | 0.106 | 0.51 | 1.000 |
| human | Catalan | Greek | 0.991 | 0.1 | -0.09 | 1.000 |
| human | Catalan | Italian | 1.085 | 0.108 | 0.82 | 1.000 |
| human | Catalan | Japanese | 0.397 | 0.045 | -8.15 | <.001 |
| human | Catalan | Russian | 0.428 | 0.048 | -7.55 | <.001 |
| human | Catalan | Spanish | 0.859 | 0.088 | -1.49 | 0.944 |
| human | Catalan | Turkish | 0.728 | 0.076 | -3.06 | 0.092 |
| human | Chinese | Dutch | 1.79 | 0.203 | 5.14 | <.001 |
| human | Chinese | English | 2.158 | 0.238 | 6.96 | <.001 |
| human | Chinese | German | 2.325 | 0.258 | 7.61 | <.001 |
| human | Chinese | Greek | 2.189 | 0.244 | 7.03 | <.001 |
| human | Chinese | Italian | 2.396 | 0.265 | 7.9 | <.001 |
| human | Chinese | Japanese | 0.877 | 0.108 | -1.07 | 0.996 |
| human | Chinese | Russian | 0.944 | 0.114 | -0.48 | 1.000 |
| human | Chinese | Spanish | 1.897 | 0.214 | 5.69 | <.001 |
| human | Chinese | Turkish | 1.606 | 0.181 | 4.2 | 0.002 |
| human | Dutch | English | 1.206 | 0.124 | 1.82 | 0.809 |
| human | Dutch | German | 1.299 | 0.134 | 2.53 | 0.322 |
| human | Dutch | Greek | 1.223 | 0.126 | 1.95 | 0.728 |
| human | Dutch | Italian | 1.339 | 0.137 | 2.84 | 0.162 |
| human | Dutch | Japanese | 0.49 | 0.057 | -6.14 | <.001 |
| human | Dutch | Russian | 0.528 | 0.06 | -5.59 | <.001 |
| human | Dutch | Spanish | 1.06 | 0.111 | 0.56 | 1.000 |
| human | Dutch | Turkish | 0.898 | 0.096 | -1.01 | 0.997 |
| human | English | German | 1.077 | 0.108 | 0.74 | 1.000 |
| human | English | Greek | 1.014 | 0.102 | 0.14 | 1.000 |
| human | English | Italian | 1.11 | 0.111 | 1.04 | 0.997 |
| human | English | Japanese | 0.406 | 0.046 | -7.95 | <.001 |
| human | English | Russian | 0.437 | 0.049 | -7.45 | <.001 |
| human | English | Spanish | 0.879 | 0.089 | -1.27 | 0.982 |
| human | English | Turkish | 0.744 | 0.077 | -2.85 | 0.158 |
| human | German | Greek | 0.942 | 0.095 | -0.6 | 1.000 |
| human | German | Italian | 1.031 | 0.103 | 0.3 | 1.000 |
| human | German | Japanese | 0.377 | 0.043 | -8.6 | <.001 |
| human | German | Russian | 0.406 | 0.045 | -8.05 | <.001 |
| human | German | Spanish | 0.816 | 0.083 | -2 | 0.690 |
| human | German | Turkish | 0.691 | 0.071 | -3.59 | 0.017 |
| human | Greek | Italian | 1.095 | 0.11 | 0.9 | 0.999 |
| human | Greek | Japanese | 0.401 | 0.046 | -8.04 | <.001 |
| human | Greek | Russian | 0.431 | 0.048 | -7.48 | <.001 |
| human | Greek | Spanish | 0.867 | 0.089 | -1.4 | 0.964 |
| human | Greek | Turkish | 0.734 | 0.077 | -2.96 | 0.119 |
| human | Italian | Japanese | 0.366 | 0.042 | -8.86 | <.001 |
| human | Italian | Russian | 0.394 | 0.044 | -8.34 | <.001 |
| human | Italian | Spanish | 0.792 | 0.08 | -2.31 | 0.471 |
| human | Italian | Turkish | 0.671 | 0.069 | -3.89 | 0.006 |
| human | Japanese | Russian | 1.077 | 0.134 | 0.6 | 1.000 |
| human | Japanese | Spanish | 2.164 | 0.251 | 6.65 | <.001 |
| human | Japanese | Turkish | 1.832 | 0.215 | 5.17 | <.001 |
| human | Russian | Spanish | 2.009 | 0.226 | 6.2 | <.001 |
| human | Russian | Turkish | 1.702 | 0.196 | 4.61 | <.001 |
| human | Spanish | Turkish | 0.847 | 0.089 | -1.58 | 0.917 |
| DS_V3 | Arabic | Catalan | 0.866 | 0.071 | -1.74 | 0.847 |
| DS_V3 | Arabic | Chinese | 1.06 | 0.087 | 0.71 | 1.000 |
| DS_V3 | Arabic | Dutch | 0.509 | 0.043 | -8.01 | <.001 |
| DS_V3 | Arabic | English | 0.953 | 0.078 | -0.59 | 1.000 |
| DS_V3 | Arabic | German | 0.384 | 0.033 | -11.17 | <.001 |
| DS_V3 | Arabic | Greek | 1.699 | 0.139 | 6.47 | <.001 |
| DS_V3 | Arabic | Italian | 0.202 | 0.018 | -17.71 | <.001 |
| DS_V3 | Arabic | Japanese | 2.727 | 0.223 | 12.29 | <.001 |
| DS_V3 | Arabic | Russian | 0.329 | 0.028 | -12.86 | <.001 |
| DS_V3 | Arabic | Spanish | 0.293 | 0.026 | -14.07 | <.001 |
| DS_V3 | Arabic | Turkish | 0.375 | 0.032 | -11.46 | <.001 |
| DS_V3 | Catalan | Chinese | 1.224 | 0.103 | 2.4 | 0.403 |
| DS_V3 | Catalan | Dutch | 0.588 | 0.051 | -6.16 | <.001 |
| DS_V3 | Catalan | English | 1.1 | 0.093 | 1.13 | 0.993 |
| DS_V3 | Catalan | German | 0.443 | 0.039 | -9.29 | <.001 |
| DS_V3 | Catalan | Greek | 1.962 | 0.164 | 8.06 | <.001 |
| DS_V3 | Catalan | Italian | 0.233 | 0.022 | -15.78 | <.001 |
| DS_V3 | Catalan | Japanese | 3.149 | 0.263 | 13.73 | <.001 |
| DS_V3 | Catalan | Russian | 0.38 | 0.034 | -10.92 | <.001 |
| DS_V3 | Catalan | Spanish | 0.338 | 0.03 | -12.16 | <.001 |
| DS_V3 | Catalan | Turkish | 0.434 | 0.038 | -9.53 | <.001 |
| DS_V3 | Chinese | Dutch | 0.48 | 0.041 | -8.55 | <.001 |
| DS_V3 | Chinese | English | 0.899 | 0.075 | -1.27 | 0.982 |
| DS_V3 | Chinese | German | 0.362 | 0.032 | -11.64 | <.001 |
| DS_V3 | Chinese | Greek | 1.602 | 0.133 | 5.66 | <.001 |
| DS_V3 | Chinese | Italian | 0.19 | 0.018 | -18.03 | <.001 |
| DS_V3 | Chinese | Japanese | 2.572 | 0.214 | 11.37 | <.001 |
| DS_V3 | Chinese | Russian | 0.31 | 0.027 | -13.27 | <.001 |
| DS_V3 | Chinese | Spanish | 0.276 | 0.025 | -14.48 | <.001 |
| DS_V3 | Chinese | Turkish | 0.354 | 0.031 | -11.89 | <.001 |
| DS_V3 | Dutch | English | 1.872 | 0.161 | 7.29 | <.001 |
| DS_V3 | Dutch | German | 0.755 | 0.067 | -3.16 | 0.069 |
| DS_V3 | Dutch | Greek | 3.338 | 0.285 | 14.13 | <.001 |
| DS_V3 | Dutch | Italian | 0.397 | 0.037 | -9.86 | <.001 |
| DS_V3 | Dutch | Japanese | 5.358 | 0.457 | 19.69 | <.001 |
| DS_V3 | Dutch | Russian | 0.646 | 0.058 | -4.84 | <.001 |
| DS_V3 | Dutch | Spanish | 0.575 | 0.052 | -6.09 | <.001 |
| DS_V3 | Dutch | Turkish | 0.738 | 0.066 | -3.41 | 0.032 |
| DS_V3 | English | German | 0.403 | 0.035 | -10.41 | <.001 |
| DS_V3 | English | Greek | 1.783 | 0.149 | 6.94 | <.001 |
| DS_V3 | English | Italian | 0.212 | 0.019 | -16.87 | <.001 |
| DS_V3 | English | Japanese | 2.862 | 0.238 | 12.65 | <.001 |
| DS_V3 | English | Russian | 0.345 | 0.03 | -12.06 | <.001 |
| DS_V3 | English | Spanish | 0.307 | 0.027 | -13.27 | <.001 |
| DS_V3 | English | Turkish | 0.394 | 0.034 | -10.66 | <.001 |
| DS_V3 | German | Greek | 4.424 | 0.383 | 17.17 | <.001 |
| DS_V3 | German | Italian | 0.526 | 0.05 | -6.77 | <.001 |
| DS_V3 | German | Japanese | 7.101 | 0.614 | 22.67 | <.001 |
| DS_V3 | German | Russian | 0.857 | 0.078 | -1.69 | 0.872 |
| DS_V3 | German | Spanish | 0.762 | 0.07 | -2.96 | 0.122 |
| DS_V3 | German | Turkish | 0.978 | 0.088 | -0.25 | 1.000 |
| DS_V3 | Greek | Italian | 0.119 | 0.011 | -23.32 | <.001 |
| DS_V3 | Greek | Japanese | 1.605 | 0.132 | 5.73 | <.001 |
| DS_V3 | Greek | Russian | 0.194 | 0.017 | -18.74 | <.001 |
| DS_V3 | Greek | Spanish | 0.172 | 0.015 | -19.9 | <.001 |
| DS_V3 | Greek | Turkish | 0.221 | 0.019 | -17.39 | <.001 |
| DS_V3 | Italian | Japanese | 13.507 | 1.234 | 28.5 | <.001 |
| DS_V3 | Italian | Russian | 1.629 | 0.156 | 5.1 | <.001 |
| DS_V3 | Italian | Spanish | 1.449 | 0.14 | 3.84 | 0.007 |
| DS_V3 | Italian | Turkish | 1.86 | 0.177 | 6.53 | <.001 |
| DS_V3 | Japanese | Russian | 0.121 | 0.011 | -24.17 | <.001 |
| DS_V3 | Japanese | Spanish | 0.107 | 0.009 | -25.28 | <.001 |
| DS_V3 | Japanese | Turkish | 0.138 | 0.012 | -22.89 | <.001 |
| DS_V3 | Russian | Spanish | 0.889 | 0.083 | -1.26 | 0.984 |
| DS_V3 | Russian | Turkish | 1.141 | 0.104 | 1.44 | 0.955 |
| DS_V3 | Spanish | Turkish | 1.284 | 0.118 | 2.71 | 0.221 |
| GPT-4o | Arabic | Catalan | 0.286 | 0.026 | -14.04 | <.001 |
| GPT-4o | Arabic | Chinese | 0.432 | 0.037 | -9.71 | <.001 |
| GPT-4o | Arabic | Dutch | 1.024 | 0.085 | 0.29 | 1.000 |
| GPT-4o | Arabic | English | 0.748 | 0.063 | -3.45 | 0.028 |
| GPT-4o | Arabic | German | 0.506 | 0.043 | -7.95 | <.001 |
| GPT-4o | Arabic | Greek | 1.364 | 0.112 | 3.77 | 0.009 |
| GPT-4o | Arabic | Italian | 0.401 | 0.035 | -10.55 | <.001 |
| GPT-4o | Arabic | Japanese | 2.375 | 0.195 | 10.51 | <.001 |
| GPT-4o | Arabic | Russian | 0.804 | 0.068 | -2.6 | 0.277 |
| GPT-4o | Arabic | Spanish | 0.211 | 0.019 | -16.87 | <.001 |
| GPT-4o | Arabic | Turkish | 0.982 | 0.081 | -0.22 | 1.000 |
| GPT-4o | Catalan | Chinese | 1.511 | 0.142 | 4.39 | <.001 |
| GPT-4o | Catalan | Dutch | 3.58 | 0.326 | 14.03 | <.001 |
| GPT-4o | Catalan | English | 2.615 | 0.24 | 10.48 | <.001 |
| GPT-4o | Catalan | German | 1.77 | 0.165 | 6.12 | <.001 |
| GPT-4o | Catalan | Greek | 4.766 | 0.43 | 17.29 | <.001 |
| GPT-4o | Catalan | Italian | 1.4 | 0.132 | 3.56 | 0.019 |
| GPT-4o | Catalan | Japanese | 8.298 | 0.746 | 23.55 | <.001 |
| GPT-4o | Catalan | Russian | 2.808 | 0.257 | 11.28 | <.001 |
| GPT-4o | Catalan | Spanish | 0.737 | 0.073 | -3.08 | 0.088 |
| GPT-4o | Catalan | Turkish | 3.432 | 0.312 | 13.55 | <.001 |
| GPT-4o | Chinese | Dutch | 2.369 | 0.209 | 9.79 | <.001 |
| GPT-4o | Chinese | English | 1.73 | 0.154 | 6.16 | <.001 |
| GPT-4o | Chinese | German | 1.171 | 0.106 | 1.74 | 0.849 |
| GPT-4o | Chinese | Greek | 3.154 | 0.276 | 13.11 | <.001 |
| GPT-4o | Chinese | Italian | 0.926 | 0.085 | -0.83 | 1.000 |
| GPT-4o | Chinese | Japanese | 5.491 | 0.478 | 19.56 | <.001 |
| GPT-4o | Chinese | Russian | 1.858 | 0.165 | 6.97 | <.001 |
| GPT-4o | Chinese | Spanish | 0.487 | 0.047 | -7.42 | <.001 |
| GPT-4o | Chinese | Turkish | 2.271 | 0.2 | 9.3 | <.001 |
| GPT-4o | Dutch | English | 0.73 | 0.063 | -3.67 | 0.013 |
| GPT-4o | Dutch | German | 0.494 | 0.043 | -8.07 | <.001 |
| GPT-4o | Dutch | Greek | 1.331 | 0.112 | 3.4 | 0.033 |
| GPT-4o | Dutch | Italian | 0.391 | 0.035 | -10.61 | <.001 |
| GPT-4o | Dutch | Japanese | 2.318 | 0.194 | 10.05 | <.001 |
| GPT-4o | Dutch | Russian | 0.784 | 0.067 | -2.84 | 0.163 |
| GPT-4o | Dutch | Spanish | 0.206 | 0.019 | -16.86 | <.001 |
| GPT-4o | Dutch | Turkish | 0.959 | 0.081 | -0.5 | 1.000 |
| GPT-4o | English | German | 0.677 | 0.06 | -4.43 | <.001 |
| GPT-4o | English | Greek | 1.823 | 0.155 | 7.05 | <.001 |
| GPT-4o | English | Italian | 0.535 | 0.048 | -6.99 | <.001 |
| GPT-4o | English | Japanese | 3.174 | 0.268 | 13.65 | <.001 |
| GPT-4o | English | Russian | 1.074 | 0.093 | 0.82 | 1.000 |
| GPT-4o | English | Spanish | 0.282 | 0.027 | -13.39 | <.001 |
| GPT-4o | English | Turkish | 1.313 | 0.113 | 3.17 | 0.067 |
| GPT-4o | German | Greek | 2.693 | 0.234 | 11.41 | <.001 |
| GPT-4o | German | Italian | 0.791 | 0.072 | -2.58 | 0.292 |
| GPT-4o | German | Japanese | 4.689 | 0.405 | 17.9 | <.001 |
| GPT-4o | German | Russian | 1.587 | 0.14 | 5.24 | <.001 |
| GPT-4o | German | Spanish | 0.416 | 0.04 | -9.12 | <.001 |
| GPT-4o | German | Turkish | 1.94 | 0.17 | 7.58 | <.001 |
| GPT-4o | Greek | Italian | 0.294 | 0.026 | -13.92 | <.001 |
| GPT-4o | Greek | Japanese | 1.741 | 0.145 | 6.68 | <.001 |
| GPT-4o | Greek | Russian | 0.589 | 0.05 | -6.23 | <.001 |
| GPT-4o | Greek | Spanish | 0.155 | 0.014 | -20.01 | <.001 |
| GPT-4o | Greek | Turkish | 0.72 | 0.061 | -3.89 | 0.006 |
| GPT-4o | Italian | Japanese | 5.929 | 0.519 | 20.34 | <.001 |
| GPT-4o | Italian | Russian | 2.006 | 0.179 | 7.79 | <.001 |
| GPT-4o | Italian | Spanish | 0.526 | 0.051 | -6.61 | <.001 |
| GPT-4o | Italian | Turkish | 2.452 | 0.217 | 10.12 | <.001 |
| GPT-4o | Japanese | Russian | 0.338 | 0.029 | -12.83 | <.001 |
| GPT-4o | Japanese | Spanish | 0.089 | 0.008 | -26.08 | <.001 |
| GPT-4o | Japanese | Turkish | 0.414 | 0.035 | -10.54 | <.001 |
| GPT-4o | Russian | Spanish | 0.262 | 0.025 | -14.18 | <.001 |
| GPT-4o | Russian | Turkish | 1.222 | 0.105 | 2.34 | 0.445 |
| GPT-4o | Spanish | Turkish | 4.659 | 0.437 | 16.4 | <.001 |
| Grok-3 | Arabic | Catalan | 0.537 | 0.046 | -7.25 | <.001 |
| Grok-3 | Arabic | Chinese | 0.599 | 0.051 | -6.03 | <.001 |
| Grok-3 | Arabic | Dutch | 0.454 | 0.04 | -9.07 | <.001 |
| Grok-3 | Arabic | English | 0.38 | 0.033 | -11.03 | <.001 |
| Grok-3 | Arabic | German | 0.805 | 0.068 | -2.57 | 0.295 |
| Grok-3 | Arabic | Greek | 1.889 | 0.156 | 7.72 | <.001 |
| Grok-3 | Arabic | Italian | 0.376 | 0.033 | -11.14 | <.001 |
| Grok-3 | Arabic | Japanese | 2.586 | 0.213 | 11.54 | <.001 |
| Grok-3 | Arabic | Russian | 0.349 | 0.031 | -11.95 | <.001 |
| Grok-3 | Arabic | Spanish | 0.336 | 0.03 | -12.23 | <.001 |
| Grok-3 | Arabic | Turkish | 0.628 | 0.054 | -5.45 | <.001 |
| Grok-3 | Catalan | Chinese | 1.115 | 0.099 | 1.22 | 0.988 |
| Grok-3 | Catalan | Dutch | 0.845 | 0.077 | -1.86 | 0.783 |
| Grok-3 | Catalan | English | 0.707 | 0.065 | -3.78 | 0.009 |
| Grok-3 | Catalan | German | 1.498 | 0.132 | 4.58 | <.001 |
| Grok-3 | Catalan | Greek | 3.516 | 0.304 | 14.56 | <.001 |
| Grok-3 | Catalan | Italian | 0.7 | 0.064 | -3.89 | 0.006 |
| Grok-3 | Catalan | Japanese | 4.813 | 0.415 | 18.23 | <.001 |
| Grok-3 | Catalan | Russian | 0.65 | 0.06 | -4.69 | <.001 |
| Grok-3 | Catalan | Spanish | 0.625 | 0.058 | -5.08 | <.001 |
| Grok-3 | Catalan | Turkish | 1.169 | 0.104 | 1.76 | 0.841 |
| Grok-3 | Chinese | Dutch | 0.757 | 0.068 | -3.08 | 0.088 |
| Grok-3 | Chinese | English | 0.634 | 0.058 | -4.99 | <.001 |
| Grok-3 | Chinese | German | 1.343 | 0.118 | 3.36 | 0.038 |
| Grok-3 | Chinese | Greek | 3.153 | 0.271 | 13.35 | <.001 |
| Grok-3 | Chinese | Italian | 0.628 | 0.057 | -5.1 | <.001 |
| Grok-3 | Chinese | Japanese | 4.316 | 0.37 | 17.05 | <.001 |
| Grok-3 | Chinese | Russian | 0.583 | 0.053 | -5.88 | <.001 |
| Grok-3 | Chinese | Spanish | 0.561 | 0.052 | -6.28 | <.001 |
| Grok-3 | Chinese | Turkish | 1.049 | 0.093 | 0.54 | 1.000 |
| Grok-3 | Dutch | English | 0.837 | 0.078 | -1.92 | 0.747 |
| Grok-3 | Dutch | German | 1.773 | 0.158 | 6.42 | <.001 |
| Grok-3 | Dutch | Greek | 4.163 | 0.364 | 16.3 | <.001 |
| Grok-3 | Dutch | Italian | 0.829 | 0.077 | -2.03 | 0.674 |
| Grok-3 | Dutch | Japanese | 5.699 | 0.498 | 19.93 | <.001 |
| Grok-3 | Dutch | Russian | 0.769 | 0.072 | -2.81 | 0.174 |
| Grok-3 | Dutch | Spanish | 0.74 | 0.069 | -3.21 | 0.059 |
| Grok-3 | Dutch | Turkish | 1.385 | 0.125 | 3.61 | 0.016 |
| Grok-3 | English | German | 2.119 | 0.191 | 8.31 | <.001 |
| Grok-3 | English | Greek | 4.974 | 0.441 | 18.11 | <.001 |
| Grok-3 | English | Italian | 0.99 | 0.093 | -0.11 | 1.000 |
| Grok-3 | English | Japanese | 6.809 | 0.601 | 21.71 | <.001 |
| Grok-3 | English | Russian | 0.919 | 0.087 | -0.9 | 0.999 |
| Grok-3 | English | Spanish | 0.884 | 0.084 | -1.3 | 0.979 |
| Grok-3 | English | Turkish | 1.654 | 0.151 | 5.52 | <.001 |
| Grok-3 | German | Greek | 2.348 | 0.199 | 10.05 | <.001 |
| Grok-3 | German | Italian | 0.467 | 0.042 | -8.43 | <.001 |
| Grok-3 | German | Japanese | 3.214 | 0.272 | 13.78 | <.001 |
| Grok-3 | German | Russian | 0.434 | 0.039 | -9.19 | <.001 |
| Grok-3 | German | Spanish | 0.417 | 0.038 | -9.58 | <.001 |
| Grok-3 | German | Turkish | 0.781 | 0.068 | -2.82 | 0.171 |
| Grok-3 | Greek | Italian | 0.199 | 0.018 | -18.22 | <.001 |
| Grok-3 | Greek | Japanese | 1.369 | 0.113 | 3.8 | 0.008 |
| Grok-3 | Greek | Russian | 0.185 | 0.016 | -18.97 | <.001 |
| Grok-3 | Greek | Spanish | 0.178 | 0.016 | -19.29 | <.001 |
| Grok-3 | Greek | Turkish | 0.333 | 0.029 | -12.83 | <.001 |
| Grok-3 | Italian | Japanese | 6.878 | 0.607 | 21.83 | <.001 |
| Grok-3 | Italian | Russian | 0.928 | 0.087 | -0.79 | 1.000 |
| Grok-3 | Italian | Spanish | 0.893 | 0.084 | -1.19 | 0.990 |
| Grok-3 | Italian | Turkish | 1.671 | 0.152 | 5.63 | <.001 |
| Grok-3 | Japanese | Russian | 0.135 | 0.012 | -22.53 | <.001 |
| Grok-3 | Japanese | Spanish | 0.13 | 0.012 | -22.83 | <.001 |
| Grok-3 | Japanese | Turkish | 0.243 | 0.021 | -16.52 | <.001 |
| Grok-3 | Russian | Spanish | 0.962 | 0.092 | -0.41 | 1.000 |
| Grok-3 | Russian | Turkish | 1.8 | 0.165 | 6.41 | <.001 |
| Grok-3 | Spanish | Turkish | 1.871 | 0.172 | 6.8 | <.001 |

**Supplementary Table 3: Pairwise comparisons of all agents per language (stability)**

| Language | Agent 1 | Agent 2 | OR | SE | z | P |
|---|---|---|---|---|---|---|
| **Arabic** | human | DS_V3 | 1.026 | 0.13 | 0.199 | 0.997 |
| | human | GPT-4o | 2.82 | 0.334 | 8.758 | < .001 |
| | human | Grok-3 | 1.015 | 0.129 | 0.114 | 0.999 |
| | DS_V3 | GPT-4o | 2.75 | 0.335 | 8.3 | < .001 |
| | DS_V3 | Grok-3 | 0.989 | 0.129 | -0.084 | 0.9998 |
| | GPT-4o | Grok-3 | 0.36 | 0.044 | -8.377 | < .001 |
| **Catalan** | human | DS_V3 | 1.277 | 0.168 | 1.86 | 0.245 |
| | human | GPT-4o | 1.117 | 0.148 | 0.832 | 0.839 |
| | human | Grok-3 | 0.808 | 0.111 | -1.544 | 0.411 |
| | DS_V3 | GPT-4o | 0.875 | 0.114 | -1.03 | 0.732 |
| | DS_V3 | Grok-3 | 0.633 | 0.085 | -3.388 | 0.004 |
| | GPT-4o | Grok-3 | 0.724 | 0.099 | -2.367 | 0.083 |
| **Chinese** | human | DS_V3 | 0.959 | 0.139 | -0.291 | 0.991 |
| | human | GPT-4o | 1.729 | 0.235 | 4.023 | < .001 |
| | human | Grok-3 | 0.922 | 0.134 | -0.557 | 0.945 |
| | DS_V3 | GPT-4o | 1.804 | 0.247 | 4.304 | < .001 |
| | DS_V3 | Grok-3 | 0.962 | 0.141 | -0.266 | 0.993 |
| | GPT-4o | Grok-3 | 0.533 | 0.074 | -4.56 | < .001 |
| **Dutch** | human | DS_V3 | 0.962 | 0.128 | -0.289 | 0.992 |
| | human | GPT-4o | 2.439 | 0.304 | 7.166 | < .001 |
| | human | Grok-3 | 0.513 | 0.075 | -4.594 | < .001 |
| | DS_V3 | GPT-4o | 2.535 | 0.317 | 7.447 | < .001 |
| | DS_V3 | Grok-3 | 0.534 | 0.078 | -4.313 | < .001 |
| | GPT-4o | Grok-3 | 0.21 | 0.029 | -11.327 | < .001 |
| **English** | human | DS_V3 | 0.403 | 0.057 | -6.402 | < .001 |
| | human | GPT-4o | 1.279 | 0.158 | 1.988 | 0.192 |
| | human | Grok-3 | 0.38 | 0.054 | -6.756 | < .001 |
| | DS_V3 | GPT-4o | 3.173 | 0.444 | 8.251 | < .001 |
| | DS_V3 | Grok-3 | 0.943 | 0.149 | -0.375 | 0.982 |
| | GPT-4o | Grok-3 | 0.297 | 0.042 | -8.582 | < .001 |
| **German** | human | DS_V3 | 0.316 | 0.055 | -6.561 | < .001 |
| | human | GPT-4o | 1.999 | 0.265 | 5.223 | < .001 |
| | human | Grok-3 | 0.186 | 0.038 | -8.263 | < .001 |
| | DS_V3 | GPT-4o | 6.33 | 1.069 | 10.926 | < .001 |
| | DS_V3 | Grok-3 | 0.59 | 0.135 | -2.305 | 0.097 |
| | GPT-4o | Grok-3 | 0.093 | 0.018 | -11.984 | < .001 |
| **Greek** | human | DS_V3 | 1.989 | 0.259 | 5.285 | < .001 |
| | human | GPT-4o | 3.062 | 0.389 | 8.798 | < .001 |
| | human | Grok-3 | 0.453 | 0.071 | -5.043 | < .001 |
| | DS_V3 | GPT-4o | 1.54 | 0.182 | 3.646 | 0.002 |
| | DS_V3 | Grok-3 | 0.228 | 0.034 | -9.863 | < .001 |
| | GPT-4o | Grok-3 | 0.148 | 0.022 | -12.96 | < .001 |
| **Italian** | human | DS_V3 | 0.252 | 0.045 | -7.797 | < .001 |
| | human | GPT-4o | 2.089 | 0.268 | 5.738 | < .001 |
| | human | Grok-3 | 0.352 | 0.057 | -6.415 | < .001 |
| | DS_V3 | GPT-4o | 8.294 | 1.419 | 12.366 | < .001 |
| | DS_V3 | Grok-3 | 1.397 | 0.277 | 1.684 | 0.332 |
| | GPT-4o | Grok-3 | 0.168 | 0.026 | -11.392 | < .001 |
| **Japanese** | human | DS_V3 | < .001 | < .001 | -3.794 | < .001 |
| | human | GPT-4o | 2.271 | 0.313 | 5.947 | < .001 |
| | human | Grok-3 | 0.492 | 0.083 | -4.184 | < .001 |
| | DS_V3 | GPT-4o | > 1,000 | > 1,000 | 3.985 | < .001 |
| | DS_V3 | Grok-3 | > 1,001 | > 1,001 | 3.624 | 0.002 |
| | GPT-4o | Grok-3 | 0.217 | 0.035 | -9.583 | < .001 |
| **Russian** | human | DS_V3 | 0.896 | 0.133 | -0.737 | 0.882 |
| | human | GPT-4o | 2.124 | 0.289 | 5.54 | < .001 |
| | human | Grok-3 | 0.651 | 0.102 | -2.744 | 0.031 |
| | DS_V3 | GPT-4o | 2.37 | 0.329 | 6.213 | < .001 |
| | DS_V3 | Grok-3 | 0.726 | 0.116 | -2.007 | 0.185 |
| | GPT-4o | Grok-3 | 0.306 | 0.045 | -8.035 | < .001 |
| **Spanish** | human | DS_V3 | 0.548 | 0.077 | -4.265 | < .001 |
| | human | GPT-4o | 0.629 | 0.087 | -3.366 | 0.004 |
| | human | Grok-3 | 0.212 | 0.036 | -9.02 | < .001 |
| | DS_V3 | GPT-4o | 1.146 | 0.17 | 0.921 | 0.794 |
| | DS_V3 | Grok-3 | 0.387 | 0.07 | -5.269 | < .001 |
| | GPT-4o | Grok-3 | 0.338 | 0.06 | -6.098 | < .001 |
| **Turkish** | human | DS_V3 | 0.382 | 0.057 | -6.486 | < .001 |
| | human | GPT-4o | 3.431 | 0.415 | 10.2 | < .001 |
| | human | Grok-3 | 0.497 | 0.071 | -4.924 | < .001 |
| | DS_V3 | GPT-4o | 8.992 | 1.269 | 15.566 | < .001 |
| | DS_V3 | Grok-3 | 1.304 | 0.208 | 1.662 | 0.344 |
| | GPT-4o | Grok-3 | 0.145 | 0.019 | -14.426 | < .001 |

**Supplementary Table 4: Pairwise comparisons of languages within each agent (stability)**

| Agent | Lang1 | Lang2 | OR | SE | z | p |
|---|---|---|---|---|---|---|
| human | Arabic | Catalan | 0.934 | 0.121 | -0.525 | 1.000 |
| human | Arabic | Chinese | 0.7 | 0.094 | -2.646 | 0.254 |
| human | Arabic | Dutch | 1.005 | 0.129 | 0.039 | 1.000 |
| human | Arabic | English | 1.292 | 0.163 | 2.038 | 0.667 |
| human | Arabic | German | 0.741 | 0.099 | -2.25 | 0.513 |
| human | Arabic | Greek | 0.819 | 0.108 | -1.519 | 0.936 |
| human | Arabic | Italian | 0.862 | 0.113 | -1.132 | 0.993 |
| human | Arabic | Japanese | 0.599 | 0.083 | -3.711 | 0.011 |
| human | Arabic | Russian | 0.648 | 0.088 | -3.2 | 0.061 |
| human | Arabic | Spanish | 1.082 | 0.138 | 0.62 | 1.000 |
| human | Arabic | Turkish | 1.127 | 0.143 | 0.937 | 0.999 |
| human | Catalan | Chinese | 0.749 | 0.104 | -2.071 | 0.643 |
| human | Catalan | Dutch | 1.076 | 0.144 | 0.547 | 1.000 |
| human | Catalan | English | 1.384 | 0.18 | 2.49 | 0.346 |
| human | Catalan | German | 0.793 | 0.109 | -1.68 | 0.878 |
| human | Catalan | Greek | 0.876 | 0.12 | -0.967 | 0.998 |
| human | Catalan | Italian | 0.923 | 0.125 | -0.589 | 1.000 |
| human | Catalan | Japanese | 0.642 | 0.091 | -3.12 | 0.078 |
| human | Catalan | Russian | 0.693 | 0.098 | -2.603 | 0.278 |
| human | Catalan | Spanish | 1.159 | 0.154 | 1.111 | 0.994 |
| human | Catalan | Turkish | 1.206 | 0.159 | 1.419 | 0.96 |
| human | Chinese | Dutch | 1.436 | 0.199 | 2.61 | 0.274 |
| human | Chinese | English | 1.846 | 0.25 | 4.53 | < .001 |
| human | Chinese | German | 1.058 | 0.151 | 0.397 | 1.000 |
| human | Chinese | Greek | 1.169 | 0.165 | 1.109 | 0.994 |
| human | Chinese | Italian | 1.232 | 0.173 | 1.488 | 0.944 |
| human | Chinese | Japanese | 0.856 | 0.126 | -1.057 | 0.996 |
| human | Chinese | Russian | 0.925 | 0.135 | -0.533 | 1.000 |
| human | Chinese | Spanish | 1.546 | 0.213 | 3.17 | 0.067 |
| human | Chinese | Turkish | 1.61 | 0.221 | 3.471 | 0.026 |
| human | Dutch | English | 1.286 | 0.167 | 1.942 | 0.733 |
| human | Dutch | German | 0.737 | 0.101 | -2.221 | 0.534 |
| human | Dutch | Greek | 0.814 | 0.111 | -1.512 | 0.938 |
| human | Dutch | Italian | 0.858 | 0.116 | -1.135 | 0.993 |
| human | Dutch | Japanese | 0.596 | 0.084 | -3.653 | 0.014 |
| human | Dutch | Russian | 0.645 | 0.09 | -3.139 | 0.074 |
| human | Dutch | Spanish | 1.077 | 0.142 | 0.563 | 1.000 |
| human | Dutch | Turkish | 1.121 | 0.147 | 0.869 | 0.999 |
| human | English | German | 0.573 | 0.077 | -4.149 | 0.002 |
| human | English | Greek | 0.633 | 0.084 | -3.45 | 0.028 |
| human | English | Italian | 0.667 | 0.088 | -3.075 | 0.088 |
| human | English | Japanese | 0.464 | 0.064 | -5.552 | < .001 |
| human | English | Russian | 0.501 | 0.069 | -5.052 | < .001 |
| human | English | Spanish | 0.837 | 0.107 | -1.382 | 0.967 |
| human | English | Turkish | 0.872 | 0.112 | -1.073 | 0.996 |
| human | German | Greek | 1.105 | 0.155 | 0.712 | 1.000 |
| human | German | Italian | 1.164 | 0.162 | 1.093 | 0.995 |
| human | German | Japanese | 0.809 | 0.118 | -1.454 | 0.953 |
| human | German | Russian | 0.874 | 0.126 | -0.932 | 0.999 |
| human | German | Spanish | 1.461 | 0.199 | 2.785 | 0.186 |
| human | German | Turkish | 1.521 | 0.206 | 3.09 | 0.085 |
| human | Greek | Italian | 1.054 | 0.145 | 0.38 | 1.000 |
| human | Greek | Japanese | 0.732 | 0.105 | -2.164 | 0.576 |
| human | Greek | Russian | 0.791 | 0.113 | -1.643 | 0.893 |
| human | Greek | Spanish | 1.322 | 0.178 | 2.076 | 0.64 |
| human | Greek | Turkish | 1.376 | 0.185 | 2.383 | 0.418 |
| human | Italian | Japanese | 0.695 | 0.1 | -2.541 | 0.314 |
| human | Italian | Russian | 0.751 | 0.106 | -2.02 | 0.68 |
| human | Italian | Spanish | 1.255 | 0.168 | 1.699 | 0.869 |
| human | Italian | Turkish | 1.306 | 0.174 | 2.006 | 0.69 |
| human | Japanese | Russian | 1.081 | 0.16 | 0.526 | 1.000 |
| human | Japanese | Spanish | 1.806 | 0.254 | 4.209 | 0.002 |
| human | Japanese | Turkish | 1.88 | 0.263 | 4.508 | < .001 |
| human | Russian | Spanish | 1.671 | 0.232 | 3.702 | 0.012 |
| human | Russian | Turkish | 1.739 | 0.241 | 3.998 | 0.004 |
| human | Spanish | Turkish | 1.041 | 0.136 | 0.308 | 1.000 |
| DS_V3 | Arabic | Catalan | 1.163 | 0.15 | 1.167 | 0.991 |
| DS_V3 | Arabic | Chinese | 0.654 | 0.09 | -3.071 | 0.089 |
| DS_V3 | Arabic | Dutch | 0.943 | 0.125 | -0.446 | 1.000 |
| DS_V3 | Arabic | English | 0.508 | 0.073 | -4.71 | < .001 |
| DS_V3 | Arabic | German | 0.228 | 0.039 | -8.63 | < .001 |
| DS_V3 | Arabic | Greek | 1.587 | 0.2 | 3.668 | 0.013 |
| DS_V3 | Arabic | Italian | 0.212 | 0.037 | -8.9 | < .001 |
| DS_V3 | Arabic | Japanese | < 0.001 | < 0.001 | -3.92 | 0.005 |
| DS_V3 | Arabic | Russian | 0.566 | 0.08 | -4.027 | 0.003 |
| DS_V3 | Arabic | Spanish | 0.579 | 0.081 | -3.884 | 0.006 |
| DS_V3 | Arabic | Turkish | 0.419 | 0.062 | -5.843 | < .001 |
| DS_V3 | Catalan | Chinese | 0.563 | 0.077 | -4.19 | 0.002 |
| DS_V3 | Catalan | Dutch | 0.811 | 0.106 | -1.601 | 0.909 |
| DS_V3 | Catalan | English | 0.437 | 0.062 | -5.794 | < .001 |
| DS_V3 | Catalan | German | 0.196 | 0.033 | -9.541 | < .001 |
| DS_V3 | Catalan | Greek | 1.365 | 0.17 | 2.491 | 0.345 |
| DS_V3 | Catalan | Italian | 0.182 | 0.032 | -9.785 | < .001 |
| DS_V3 | Catalan | Japanese | < 0.001 | < 0.001 | -3.954 | 0.004 |
| DS_V3 | Catalan | Russian | 0.487 | 0.068 | -5.131 | < .001 |
| DS_V3 | Catalan | Spanish | 0.498 | 0.07 | -4.987 | < .001 |
| DS_V3 | Catalan | Turkish | 0.36 | 0.053 | -6.892 | < .001 |
| DS_V3 | Chinese | Dutch | 1.441 | 0.202 | 2.613 | 0.272 |
| DS_V3 | Chinese | English | 0.776 | 0.117 | -1.678 | 0.879 |
| DS_V3 | Chinese | German | 0.349 | 0.062 | -5.933 | < .001 |
| DS_V3 | Chinese | Greek | 2.426 | 0.326 | 6.603 | < .001 |
| DS_V3 | Chinese | Italian | 0.324 | 0.059 | -6.24 | < .001 |
| DS_V3 | Chinese | Japanese | < 0.001 | < 0.001 | -3.819 | 0.007 |
| DS_V3 | Chinese | Russian | 0.865 | 0.129 | -0.975 | 0.998 |
| DS_V3 | Chinese | Spanish | 0.885 | 0.131 | -0.827 | 1.000 |
| DS_V3 | Chinese | Turkish | 0.641 | 0.1 | -2.855 | 0.158 |
| DS_V3 | Dutch | English | 0.539 | 0.078 | -4.254 | 0.001 |
| DS_V3 | Dutch | German | 0.242 | 0.042 | -8.211 | < .001 |
| DS_V3 | Dutch | Greek | 1.683 | 0.215 | 4.075 | 0.003 |
| DS_V3 | Dutch | Italian | 0.225 | 0.04 | -8.476 | < .001 |
| DS_V3 | Dutch | Japanese | < 0.001 | < 0.001 | -3.905 | 0.005 |
| DS_V3 | Dutch | Russian | 0.6 | 0.086 | -3.571 | 0.018 |
| DS_V3 | Dutch | Spanish | 0.614 | 0.087 | -3.425 | 0.03 |
| DS_V3 | Dutch | Turkish | 0.445 | 0.067 | -5.387 | < .001 |
| DS_V3 | English | German | 0.449 | 0.082 | -4.395 | < .001 |
| DS_V3 | English | Greek | 3.126 | 0.438 | 8.137 | < .001 |
| DS_V3 | English | Italian | 0.417 | 0.077 | -4.72 | < .001 |
| DS_V3 | English | Japanese | < 0.001 | < 0.001 | -3.759 | 0.009 |
| DS_V3 | English | Russian | 1.115 | 0.172 | 0.705 | 1.000 |
| DS_V3 | English | Spanish | 1.14 | 0.175 | 0.854 | 0.999 |
| DS_V3 | English | Turkish | 0.826 | 0.133 | -1.191 | 0.99 |
| DS_V3 | German | Greek | 6.956 | 1.171 | 11.526 | < .001 |
| DS_V3 | German | Italian | 0.928 | 0.193 | -0.36 | 1.000 |
| DS_V3 | German | Japanese | < 0.001 | < 0.001 | -3.57 | 0.018 |
| DS_V3 | German | Russian | 2.481 | 0.447 | 5.047 | < .001 |
| DS_V3 | German | Spanish | 2.537 | 0.456 | 5.182 | < .001 |
| DS_V3 | German | Turkish | 1.837 | 0.342 | 3.269 | 0.05 |
| DS_V3 | Greek | Italian | 0.133 | 0.023 | -11.733 | < .001 |
| DS_V3 | Greek | Japanese | < 0.001 | < 0.001 | -4.027 | 0.003 |
| DS_V3 | Greek | Russian | 0.357 | 0.049 | -7.504 | < .001 |
| DS_V3 | Greek | Spanish | 0.365 | 0.05 | -7.369 | < .001 |
| DS_V3 | Greek | Turkish | 0.264 | 0.038 | -9.165 | < .001 |
| DS_V3 | Italian | Japanese | < 0.001 | < 0.001 | -3.553 | 0.02 |
| DS_V3 | Italian | Russian | 2.673 | 0.49 | 5.366 | < .001 |
| DS_V3 | Italian | Spanish | 2.734 | 0.5 | 5.5 | < .001 |
| DS_V3 | Italian | Turkish | 1.979 | 0.374 | 3.61 | 0.016 |
| DS_V3 | Japanese | Russian | > 1,000 | > 1,000 | 3.784 | 0.009 |
| DS_V3 | Japanese | Spanish | > 1,000 | > 1,000 | 3.789 | 0.008 |
| DS_V3 | Japanese | Turkish | > 1,000 | > 1,000 | 3.714 | 0.011 |
| DS_V3 | Russian | Spanish | 1.023 | 0.155 | 0.149 | 1.000 |
| DS_V3 | Russian | Turkish | 0.741 | 0.118 | -1.892 | 0.765 |
| DS_V3 | Spanish | Turkish | 0.724 | 0.115 | -2.039 | 0.666 |
| GPT-4o | Arabic | Catalan | 0.37 | 0.045 | -8.123 | < .001 |
| GPT-4o | Arabic | Chinese | 0.429 | 0.052 | -7.025 | < .001 |
| GPT-4o | Arabic | Dutch | 0.869 | 0.099 | -1.229 | 0.987 |
| GPT-4o | Arabic | English | 0.586 | 0.069 | -4.567 | < .001 |
| GPT-4o | Arabic | German | 0.525 | 0.062 | -5.454 | < .001 |
| GPT-4o | Arabic | Greek | 0.889 | 0.101 | -1.037 | 0.997 |
| GPT-4o | Arabic | Italian | 0.639 | 0.074 | -3.859 | 0.006 |
| GPT-4o | Arabic | Japanese | 0.482 | 0.057 | -6.123 | < .001 |
| GPT-4o | Arabic | Russian | 0.488 | 0.058 | -6.04 | < .001 |
| GPT-4o | Arabic | Spanish | 0.241 | 0.031 | -10.939 | < .001 |
| GPT-4o | Arabic | Turkish | 1.371 | 0.153 | 2.819 | 0.172 |
| GPT-4o | Catalan | Chinese | 1.16 | 0.15 | 1.145 | 0.993 |
| GPT-4o | Catalan | Dutch | 2.35 | 0.291 | 6.905 | < .001 |
| GPT-4o | Catalan | English | 1.584 | 0.2 | 3.636 | 0.015 |
| GPT-4o | Catalan | German | 1.42 | 0.181 | 2.749 | 0.203 |
| GPT-4o | Catalan | Greek | 2.402 | 0.297 | 7.09 | < .001 |
| GPT-4o | Catalan | Italian | 1.727 | 0.217 | 4.342 | < .001 |
| GPT-4o | Catalan | Japanese | 1.304 | 0.167 | 2.069 | 0.645 |
| GPT-4o | Catalan | Russian | 1.319 | 0.169 | 2.158 | 0.581 |
| GPT-4o | Catalan | Spanish | 0.652 | 0.09 | -3.085 | 0.086 |
| GPT-4o | Catalan | Turkish | 3.705 | 0.451 | 10.752 | < .001 |
| GPT-4o | Chinese | Dutch | 2.026 | 0.247 | 5.798 | < .001 |
| GPT-4o | Chinese | English | 1.365 | 0.17 | 2.5 | 0.34 |
| GPT-4o | Chinese | German | 1.224 | 0.154 | 1.608 | 0.907 |
| GPT-4o | Chinese | Greek | 2.071 | 0.252 | 5.986 | < .001 |
| GPT-4o | Chinese | Italian | 1.488 | 0.184 | 3.211 | 0.06 |
| GPT-4o | Chinese | Japanese | 1.124 | 0.142 | 0.926 | 0.999 |
| GPT-4o | Chinese | Russian | 1.137 | 0.144 | 1.015 | 0.997 |
| GPT-4o | Chinese | Spanish | 0.562 | 0.077 | -4.211 | 0.002 |
| GPT-4o | Chinese | Turkish | 3.194 | 0.383 | 9.695 | < .001 |
| GPT-4o | Dutch | English | 0.674 | 0.08 | -3.333 | 0.041 |
| GPT-4o | Dutch | German | 0.604 | 0.072 | -4.22 | 0.001 |
| GPT-4o | Dutch | Greek | 1.022 | 0.118 | 0.191 | 1.000 |
| GPT-4o | Dutch | Italian | 0.735 | 0.086 | -2.622 | 0.267 |
| GPT-4o | Dutch | Japanese | 0.555 | 0.067 | -4.892 | < .001 |
| GPT-4o | Dutch | Russian | 0.561 | 0.067 | -4.805 | < .001 |
| GPT-4o | Dutch | Spanish | 0.277 | 0.036 | -9.773 | < .001 |
| GPT-4o | Dutch | Turkish | 1.577 | 0.179 | 4.02 | 0.003 |
| GPT-4o | English | German | 0.896 | 0.11 | -0.895 | 0.999 |
| GPT-4o | English | Greek | 1.517 | 0.179 | 3.524 | 0.022 |
| GPT-4o | English | Italian | 1.09 | 0.131 | 0.715 | 1.000 |
| GPT-4o | English | Japanese | 0.823 | 0.101 | -1.577 | 0.918 |
| GPT-4o | English | Russian | 0.833 | 0.102 | -1.489 | 0.944 |
| GPT-4o | English | Spanish | 0.412 | 0.055 | -6.633 | < .001 |
| GPT-4o | English | Turkish | 2.339 | 0.272 | 7.305 | < .001 |
| GPT-4o | German | Greek | 1.692 | 0.202 | 4.409 | < .001 |
| GPT-4o | German | Italian | 1.216 | 0.148 | 1.609 | 0.906 |
| GPT-4o | German | Japanese | 0.919 | 0.114 | -0.683 | 1.000 |
| GPT-4o | German | Russian | 0.929 | 0.115 | -0.595 | 1.000 |
| GPT-4o | German | Spanish | 0.459 | 0.062 | -5.775 | < .001 |
| GPT-4o | German | Turkish | 2.61 | 0.306 | 8.17 | < .001 |
| GPT-4o | Greek | Italian | 0.719 | 0.084 | -2.812 | 0.175 |
| GPT-4o | Greek | Japanese | 0.543 | 0.065 | -5.081 | < .001 |
| GPT-4o | Greek | Russian | 0.549 | 0.066 | -4.994 | < .001 |
| GPT-4o | Greek | Spanish | 0.271 | 0.036 | -9.948 | < .001 |
| GPT-4o | Greek | Turkish | 1.542 | 0.174 | 3.83 | 0.007 |
| GPT-4o | Italian | Japanese | 0.755 | 0.093 | -2.29 | 0.484 |
| GPT-4o | Italian | Russian | 0.764 | 0.093 | -2.202 | 0.548 |
| GPT-4o | Italian | Spanish | 0.378 | 0.05 | -7.313 | < .001 |
| GPT-4o | Italian | Turkish | 2.146 | 0.248 | 6.608 | < .001 |
| GPT-4o | Japanese | Russian | 1.011 | 0.126 | 0.089 | 1.000 |
| GPT-4o | Japanese | Spanish | 0.5 | 0.068 | -5.112 | < .001 |
| GPT-4o | Japanese | Turkish | 2.841 | 0.336 | 8.821 | < .001 |
| GPT-4o | Russian | Spanish | 0.494 | 0.067 | -5.2 | < .001 |
| GPT-4o | Russian | Turkish | 2.81 | 0.332 | 8.738 | < .001 |
| GPT-4o | Spanish | Turkish | 5.682 | 0.735 | 13.426 | < .001 |
| Grok-3 | Arabic | Catalan | 0.744 | 0.101 | -2.173 | 0.57 |
| Grok-3 | Arabic | Chinese | 0.636 | 0.089 | -3.251 | 0.053 |
| Grok-3 | Arabic | Dutch | 0.509 | 0.073 | -4.7 | < .001 |
| Grok-3 | Arabic | English | 0.484 | 0.07 | -4.995 | < .001 |
| Grok-3 | Arabic | German | 0.136 | 0.027 | -9.989 | < .001 |
| Grok-3 | Arabic | Greek | 0.366 | 0.056 | -6.551 | < .001 |
| Grok-3 | Arabic | Italian | 0.299 | 0.048 | -7.528 | < .001 |
| Grok-3 | Arabic | Japanese | 0.291 | 0.047 | -7.65 | < .001 |
| Grok-3 | Arabic | Russian | 0.416 | 0.062 | -5.877 | < .001 |
| Grok-3 | Arabic | Spanish | 0.227 | 0.039 | -8.646 | < .001 |
| Grok-3 | Arabic | Turkish | 0.552 | 0.079 | -4.174 | 0.002 |
| Grok-3 | Catalan | Chinese | 0.855 | 0.123 | -1.086 | 0.995 |
| Grok-3 | Catalan | Dutch | 0.683 | 0.102 | -2.551 | 0.308 |
| Grok-3 | Catalan | English | 0.65 | 0.098 | -2.862 | 0.155 |
| Grok-3 | Catalan | German | 0.183 | 0.037 | -8.332 | < .001 |
| Grok-3 | Catalan | Greek | 0.491 | 0.078 | -4.485 | < .001 |
| Grok-3 | Catalan | Italian | 0.402 | 0.066 | -5.52 | < .001 |
| Grok-3 | Catalan | Japanese | 0.39 | 0.065 | -5.657 | < .001 |
| Grok-3 | Catalan | Russian | 0.558 | 0.086 | -3.772 | 0.009 |
| Grok-3 | Catalan | Spanish | 0.304 | 0.054 | -6.742 | < .001 |
| Grok-3 | Catalan | Turkish | 0.742 | 0.109 | -2.023 | 0.677 |
| Grok-3 | Chinese | Dutch | 0.799 | 0.122 | -1.473 | 0.948 |
| Grok-3 | Chinese | English | 0.76 | 0.117 | -1.787 | 0.826 |
| Grok-3 | Chinese | German | 0.214 | 0.044 | -7.488 | < .001 |
| Grok-3 | Chinese | Greek | 0.575 | 0.093 | -3.439 | 0.029 |
| Grok-3 | Chinese | Italian | 0.47 | 0.079 | -4.501 | < .001 |
| Grok-3 | Chinese | Japanese | 0.457 | 0.077 | -4.643 | < .001 |
| Grok-3 | Chinese | Russian | 0.653 | 0.103 | -2.708 | 0.222 |
| Grok-3 | Chinese | Spanish | 0.356 | 0.064 | -5.775 | < .001 |
| Grok-3 | Chinese | Turkish | 0.868 | 0.13 | -0.94 | 0.999 |
| Grok-3 | Dutch | English | 0.951 | 0.15 | -0.316 | 1.000 |
| Grok-3 | Dutch | German | 0.267 | 0.056 | -6.296 | < .001 |
| Grok-3 | Dutch | Greek | 0.719 | 0.119 | -1.994 | 0.698 |
| Grok-3 | Dutch | Italian | 0.588 | 0.101 | -3.089 | 0.085 |
| Grok-3 | Dutch | Japanese | 0.571 | 0.099 | -3.237 | 0.055 |
| Grok-3 | Dutch | Russian | 0.817 | 0.132 | -1.248 | 0.985 |
| Grok-3 | Dutch | Spanish | 0.445 | 0.081 | -4.426 | < .001 |
| Grok-3 | Dutch | Turkish | 1.086 | 0.168 | 0.533 | 1.000 |
| Grok-3 | English | German | 0.281 | 0.059 | -6.033 | < .001 |
| Grok-3 | English | Greek | 0.756 | 0.126 | -1.682 | 0.877 |
| Grok-3 | English | Italian | 0.618 | 0.107 | -2.781 | 0.188 |
| Grok-3 | English | Japanese | 0.601 | 0.105 | -2.93 | 0.13 |
| Grok-3 | English | Russian | 0.859 | 0.14 | -0.933 | 0.999 |
| Grok-3 | English | Spanish | 0.468 | 0.086 | -4.131 | 0.002 |
| Grok-3 | English | Turkish | 1.142 | 0.178 | 0.849 | 1.000 |
| Grok-3 | German | Greek | 2.689 | 0.581 | 4.577 | < .001 |
| Grok-3 | German | Italian | 2.199 | 0.486 | 3.562 | 0.019 |
| Grok-3 | German | Japanese | 2.136 | 0.474 | 3.419 | 0.031 |
| Grok-3 | German | Russian | 3.056 | 0.652 | 5.239 | < .001 |
| Grok-3 | German | Spanish | 1.666 | 0.383 | 2.222 | 0.533 |
| Grok-3 | German | Turkish | 4.062 | 0.846 | 6.731 | < .001 |
| Grok-3 | Greek | Italian | 0.818 | 0.147 | -1.117 | 0.994 |
| Grok-3 | Greek | Japanese | 0.795 | 0.144 | -1.271 | 0.983 |
| Grok-3 | Greek | Russian | 1.137 | 0.193 | 0.753 | 1.000 |
| Grok-3 | Greek | Spanish | 0.62 | 0.118 | -2.516 | 0.33 |
| Grok-3 | Greek | Turkish | 1.511 | 0.247 | 2.519 | 0.328 |
| Grok-3 | Italian | Japanese | 0.972 | 0.182 | -0.154 | 1.000 |
| Grok-3 | Italian | Russian | 1.39 | 0.245 | 1.866 | 0.781 |
| Grok-3 | Italian | Spanish | 0.758 | 0.148 | -1.416 | 0.961 |
| Grok-3 | Italian | Turkish | 1.847 | 0.315 | 3.604 | 0.016 |
| Grok-3 | Japanese | Russian | 1.43 | 0.254 | 2.017 | 0.682 |
| Grok-3 | Japanese | Spanish | 0.78 | 0.153 | -1.264 | 0.983 |
| Grok-3 | Japanese | Turkish | 1.901 | 0.326 | 3.748 | 0.01 |
| Grok-3 | Russian | Spanish | 0.545 | 0.102 | -3.245 | 0.054 |
| Grok-3 | Russian | Turkish | 1.329 | 0.213 | 1.778 | 0.83 |
| Grok-3 | Spanish | Turkish | 2.438 | 0.442 | 4.916 | < .001 |